\documentclass[10pt,twocolumn,letterpaper]{article}

\usepackage[pagenumbers]{wacv} 
\usepackage[accsupp]{axessibility}
\usepackage{graphicx}
\usepackage{booktabs}
\usepackage{array}
\usepackage{multirow}
\usepackage{tabularx}
\usepackage{makecell}
\usepackage{tikz}
\usetikzlibrary{spy}
\usetikzlibrary{calc}
\usepackage{xcolor}
\definecolor{GHOST}{HTML}{FFFFFF}

\usepackage[normalem]{ulem}
\useunder{\uline}{\ul}{}

\definecolor{wacvblue}{rgb}{0.21,0.49,0.74}
\usepackage[pagebackref,breaklinks,colorlinks,allcolors=wacvblue]{hyperref}

\def\wacvPaperID{885} 
\def\confName{WACV}
\def\confYear{2026}

\usepackage{xcolor}

\newcommand{\revise}[1]{\textcolor{black}{#1}}

\title{Unsupervised Segmentation by Diffusing, Walking and Cutting}

\author{Daniela Ivanova\\
University of Glasgow,\\
UK\\
\and
Marco Aversa\\
Independent,\\
Switzerland\\
\and
Paul Henderson\\
University of Glasgow,\\
UK\\
\and
John Williamson\\
University of Glasgow,\\
UK\\
}

\begin{document}

\twocolumn[{
\renewcommand\twocolumn[1][]{#1} 

\maketitle 
\centering
\vspace{-1.0em}
\includegraphics[width=0.98\linewidth]{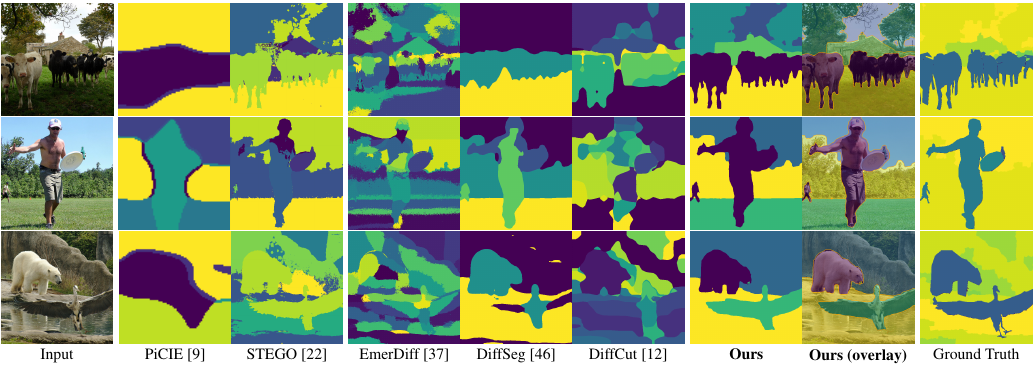}
\vspace{-1.3em}
\captionof{figure}{We perform unsupervised segmentation by applying Normalised Cuts~\cite{shi2000normalized} to self-attention features from Stable Diffusion~\cite{rombach22sd} in a hyperparameter-free setting, and achieve superior performance on Coco-Stuff-27~\cite{caesar2018coco} compared to trained and zero-shot methods.
\vspace{1em}}
\label{fig:qualitative-small} 
}]

\begin{abstract}
We propose a zero-shot unsupervised image segmentation method by utilising self-attention activations extracted from Stable Diffusion. We demonstrate that self-attention can directly be interpreted as transition probabilities in a Markov random walk between image patches. This property enables us to modulate multi-hop relationships through matrix exponentiation, which captures k-step transitions between patches. We then construct a graph representation based on self-attention feature similarity and apply Normalised Cuts to cluster them. We quantitatively analyse the effects of incorporating multi-node paths when constructing the NCuts adjacency matrix, showing that higher-order transitions enhance hierarchical relationships in the proposed segmentations. Finally, we describe an approach to automatically determine the NCut threshold criterion, avoiding the need to manually tune it. Our approach surpasses all existing methods for zero-shot unsupervised segmentation based on pre-trained diffusion models features, achieving state-of-the-art results on COCO-Stuff-27, Cityscapes and ADE20K.
\end{abstract}

\section{Introduction}
\label{sec:introduction}

Segmentation of images into semantically-meaningful regions is of key interest in computer vision. \textit{Unsupervised} \revise{semantic} segmentation aims to label images without explicit training to identify specific object types~\cite{shi2000normalized, ji2019invariant, hamilton2022unsupervised}. \revise{Zero-shot segmentation eschews training altogether, repurposing models which have not been trained for a segmentation task at all, with the goal of improved generalization~\cite{tang2023emergent, tian2024diffuse, couairon2024zero, karmann2025repurposing}.} We focus on \textit{unsupervised \revise{zero-shot} multi-class semantic segmentation}, where an image is labelled with many distinct and interpretable labels (``horse, sand, human, sky'') rather than e.g.~binary partitioning into foreground/background as in~\cite{melas2022deep, wang2023tokencut, karmann2025repurposing}.

Modern approaches to unsupervised segmentation leverage the information inherently captured in the latent features of models trained for other tasks. \revise{While some methods require access to the target data for downstream unsupervised segmentation training~\cite{caron2021emerging, hamilton2022unsupervised, Niu_2024_CVPR, Hahn:2025:UPS, Sick_2024_CVPR, 2024eagle}, and hence, cannot segment outside of their target domain, zero-shot approaches solve this by directly utilizing the features of generative text-to-image models like Stable Diffusion~\cite{rombach22sd} for semantic~\cite{tang2023emergent, tian2024diffuse, couairon2024zero} or even interactive open-vocabulary segmentation~\cite{karmann2025repurposing}.}
\revise{Stable Diffusion's self-attention features richly encode} how much each (latent) pixel attends to every other pixel, and form a natural basis for partitioning an image~\cite{tian2024diffuse}. Because the self-supervised training process needs to parsimoniously account for e.g.~spatial equivariance of rigid objects or deformation of flexible objects like human bodies, the latent features would be expected to give rise to similar embeddings for pixels across objects that are consistent under these types of transformations.

Leveraging the effectiveness of self-attention features from Stable Diffusion 1.4, as in DiffSeg~\cite{tian2024diffuse}, we propose a novel unsupervised \revise{zero-shot} segmentation approach. Departing from DiffSeg's bottom-up, iterative merging strategy, we employ the classic Normalised Cuts algorithm~\cite{shi2000normalized} to iteratively partition image regions based on self-attention similarity. To address the challenge of manual tuning in Normalised Cuts, we propose a method that automatically determines the stopping criterion. This yields a tuning-free method that automatically estimates optimal thresholds for halting the iterative partitioning, surpassing the performance of existing zero-shot unsupervised segmentation techniques.

Furthermore, we interpret the self-attention maps as transition kernels for a random walk across the image space, effectively grouping latent pixels with high connectivity probability. By exponentiating these kernels, we enable random walks to take larger steps, providing a natural and controllable mechanism for selecting the desired semantic granularity of the segmentation. This mathematical interpretation not only provides theoretical grounding for our approach but also enables intuitive control over segmentation detail.

Our contributions are:
\begin{itemize}
\item A zero-shot unsupervised semantic segmentation algorithm that outperforms the state-of-the-art, including DiffSeg~\cite{tian2024diffuse}, EmerDiff~\cite{namekata2024emerdiff}, and DiffCut~\cite{couairon2024zero};
\item A random walk interpretation of self-attention features used to build the Normalised Cuts adjacency matrix, which can modulate semantic granularity by simple matrix exponentiation;
\item A novel approach to \textit{automatically} selecting the critical threshold for Normalised Cuts on a per-image basis that eliminates the main ``fiddle factor'' in unsupervised segmentation.
\end{itemize}

\section{Related Work}
\label{sec:related_works}
\subsection{Unsupervised Segmentation via Generative Models}

Internal features of GANs \cite{goodfellow14nips}, VAEs \cite{kingma2013auto} and other self-supervised models such as DINO \cite{amir2021deep, caron2021emerging}, have long been used for diverse downstream image tasks, achieving impressive results~\cite{collins2020editing, li2021semantic, tumanyan2022splicing, hamilton2022unsupervised, tritrong2021repurposing, li2023mask, chen2023generative}.
Recently, large-scale generative text-to-image models (e.g.~Stable Diffusion \cite{rombach22sd}), have provided even richer internal features to be used for downstream tasks including depth estimation~\cite{chen2306beyond}, semantic correspondence~\cite{luo2023dhf, tang2023emergent, zhang2024tale, hedlin2024unsupervised}, and segmentation~\cite{baranchuk2021label, wu2023diffumask}. Unsupervised zero-shot segmentation is one example where methods leveraging diffusion model features have achieved state-of-the-art results \cite{namekata2024emerdiff,tian2024diffuse, couairon2024zero}.
\textbf{EmerDiff}~\cite{namekata2024emerdiff} demonstrates that cross-attention query vectors encode semantic information at each spatial location. They first perform k-means clustering on these vectors at low resolution to obtain initial segments. To achieve high-resolution segmentation, they perturb the cross-attention activations corresponding to each segment, re-generate, and observe which pixels in the output are most affected. Although this method produces accurate segmentation boundaries, it requires two denoising passes through the diffusion model \emph{per segment}. Additionally, the approach requires prior knowledge of the number of segments, meaning it is not truly unsupervised.

\textbf{DiffSeg}~\cite{tian2024diffuse} instead uses self-attention maps from Stable Diffusion \cite{rombach22sd}. For an input image yielding N latent patches, they extract an N×N self-attention matrix where each row represents the attention distribution from one patch over all other patches. Inspired by SAM \cite{kirillov2023segment}, they sample ``anchor" points on a grid and iteratively merge regions based on the Kullback-Leibler divergence between their averaged attention distributions. This bottom-up approach does not require the number of segments to be provided \textit{a priori}. However, it is computationally expensive due to both the grid sampling process and KL divergence calculations between large matrices; the latter must be calculated twice between each pair of patches/segments as it is not symmetric. Furthermore, the grid sampling strategy may miss small but significant image features if the anchors aren't sampled densely enough, and the iterative merging requires pre-setting a threshold over KLD as well the number of iterative merging passes.

\textbf{DiffCut}~\cite{couairon2024zero}, inspired by DiffSeg, extracts features (rather than raw self-attentions) from a Stable Diffusion XL model, and employs Normalised Cuts~\cite{shi2000normalized} iteratively with a manually tuned stopping criterion, and an exponentiation factor to increase the contrast of the adjacency matrix for easier partitioning. This follows from a line of recent works which has revitalized the use of spectral clustering methods by leveraging graphs constructed from neural network features. Deep Spectral Methods~\cite{melas2022deep}, for instance, utilize features extracted from a pre-trained self-supervised vision transformer (DINO) for salient object localization. Similarly, TokenCut~\cite{wang2023tokencut} and MaskCut~\cite{wang2023cut} employ Normalised Cuts on DINO features, with MaskCut iteratively partitioning the image into a pre-defined number of segments. More recently, Panoptic Cut~\cite{kang2024lazy} extends this approach by applying Normalised Cuts for a fixed number of iterations on DINO features, generating segment proposals for open vocabulary segmentation. \revise{Concurrent works interpret vision transformer attention activations as transition probabilities~\cite{el2025towards, erel2025attention, karmann2025repurposing}}. Notably, \cite{zhang2024deciphering} first explored the effectiveness of spectral clustering over features from generative diffusion models, demonstrating they encode rich semantic and spatial information. 

A common limitation across all the aforementioned segmentation methods is their reliance on either a single bipartition for salient object detection (Deep Spectral Methods, TokenCut) or iterative partitioning with manually specified parameters like the number of clusters (MaskCut), partitioning iterations (PanopticCut) or stopping criteria (DiffCut). Instead, our approach extracts and aggregates self-attention activations from Stable Diffusion 1.4 as in DiffSeg, to which we apply Normalised Cuts, concurrently to DiffCut. In addition, we propose a way to automatically halt the partitioning of each image, eliminating the need for manual tuning of NCut parameters, such as number of iterations or stopping threshold. Finally, we observe a special property of self-attention activations in particular, which allows us to control the partitioning for semantic granularity. Next, we review the theory behind the classic Normalised Cuts algorithm which informed our approach.

\subsection{Normalised cuts}
\label{subsec:relwork-spectral-clustering}

Spectral graph theory traces its roots to the 1970s~\cite{donath1973lower,fiedler1973algebraic}, but was first famously applied to computer vision by Shi and Malik~\cite{shi2000normalized} with the advent of Normalised Cuts. 

Here, an image is represented as a weighted, undirected graph $G=(V,E,w)$ where pixels correspond to vertices \revise{$V$}, and edge \revise{$E$} weights \revise{$w$} represent some similarity measure between each pair of pixels. The graph can be partitioned, or cut, into two disjoint sets $A$ and $B$, where the cost of cutting it is the sum of all edge weights that are removed:
\newcommand{\cut}{\operatorname{cut}}
\newcommand{\ncut}{\operatorname{NCut}}
\newcommand{\assoc}{\operatorname{assoc}}
\begin{equation}
    \cut(A, B) = \sum_{u \in A, v \in B} w(u, v).
    \label{eq:cut}
\end{equation}
The normalised cut (NCut) instead normalizes this cost relative to the total connections in each part of the graph:
\begin{equation}
\ncut(A, B) = \frac{\cut(A, B)}{\assoc(A, V)} + \frac{\cut(A, B)}{\assoc(B, V)}
\label{eq:ncut}
\end{equation}
%
Here \revise{$\assoc(A,V)$ measures the total connection (weighted sum of all edges) from nodes in $A$ to all nodes.} This normalization prevents the algorithm from isolating individual vertices, since cutting off a small set would result in a high NCut value. Instead the minimum-cost cut tends to split the image into two large but dissimilar regions.

The authors show that the normalised cut formulation can be solved through a generalized eigenvalue problem. Given a adjacency matrix $A$ (representing the weighted graph $G$) containing edge weights between nodes and its corresponding degree matrix $D$ (where each diagonal entry is the sum of that row in $A$), the NCut optimisation is equivalent to solving:
\begin{equation}
    Lx = \lambda Dx, 
    \label{eq:L_eigen}
\end{equation}
where $L$ is the Laplacian $L=D-A$. The solution is found in the eigenvector $x_2$ corresponding to the second-smallest eigenvalue $\lambda_2$, also known as the Fiedler vector, which is used to describe a graph's algebraic connectivity~\cite{fiedler1973algebraic}. This eigenvector assigns a value to each pixel; the algorithm finds a threshold in these values that minimizes the Normalised Cut, bipartitioning the image. Since the Fiedler vector takes on continuous values, the authors recommend testing several thresholds, and selecting the one with the smallest NCut cost. They also extend the algorithm to give more than two regions, by iteratively partitioning each resulting subgraphs until some condition, such as a target number of clusters, or a threshold NCut cost, is met.
\vspace{-12pt}
\paragraph{Dynamic NCut threshold.}
A further extension to the NCuts algorithm is the Minimum Average Node Cut (MAN-C) \cite{wang2007graph}, which aims to overcome the need for manually setting the number of clusters, iterations, or threshold NCut cost. It defines a ``good'' cluster based on the condition $\text{min-cut}(G) \geq \frac{nT}{2m}$, where $n$ is the number of vertices, $T$ is the total edge weight, and $m$ is the number of edges. In MAN-C, the \textit{minimum cut} for a partition is the smallest sum of edge weights needed to separate the graph into two parts. For each proposed Normalised Cuts partition, MAN-C checks if this minimum cut is lower than the threshold $\frac{nT}{2m}$ to determine when to halt the recursive partitioning.
\vspace{-12pt}
\paragraph{Random walk interpretation.}
Meila and Shi~\cite{NIPS2000_069654d5} later gave a probabilistic interpretation of NCut  by showing that spectral clustering can be viewed through the lens of random walks on graphs. They demonstrated that the degree-normalised adjacency matrix $P=D^{-1}A$ represents transition probabilities of a random walk, where $P_{i,j}$ represents the probability of moving from node $i$ to node $j$. The eigenvectors of this transition matrix correspond exactly to the NCut solution, with the second largest eigenvector providing the optimal binary partition. This interpretation reveals that spectral clustering effectively groups nodes that have high probability of being connected by random walks and suggests that any matrix with random walk properties (row-stochastic, capturing similarity relationships) can be directly used in the NCut framework.

\vspace{-5pt}
\section{Methods}
\label{sec:methods}

The success of DiffSeg~\cite{tian2024diffuse} and EmerDiff~\cite{tang2023emergent} demonstrates that pre-trained text-to-image diffusion models encode rich semantic relationships in their internal representations. Our approach recognizes and exploits a fundamental mathematical property of these models' self-attention matrices. In this section, we present our method in five parts: first, we describe how we extract self-attention features from Stable Diffusion (Section~\ref{subsec:method-feature-extraction}); next, we explore two complementary approaches to leveraging these features---directly interpreting self-attention as random walks (Section~\ref{subsec:method-attention-random-walk}) and constructing adjacency matrices from attention feature similarity (Section~\ref{subsec:method-feature-adjacency}); we then detail our application of Normalized Cuts to these representations (Section~\ref{subsec:method-ncut}); propose an automatic threshold determination method (Section~\ref{subsec:method-automatic-thresh}); and finally describe our upsampling process (Section~\ref{subsec:method-mask-upsampling}).

\subsection{Extracting diffusion features}
\label{subsec:method-feature-extraction}

Given the pre-trained diffusion model, we use DDIM Inversion~\cite{song2021ddim, dhariwal2021diffusion} to invert real images for a small number of steps (we use 10 out of 50 in our experiments, and ablate this in the Supplementary), then extract self-attention from the denoising U-Net decoder from a single generation step. For Stable Diffusion, transformer activations exist in four resolutions across the decoder: $8^2$, $16^2$, $32^2$ and $64^2$. We extract self-attention maps $S^{(r)} \in \mathbb{R}^{N \times N}$ where $r\in\{0,1,2,3\}$ is the resolution level and $N$ corresponds to the spatially flattened latent feature resolution $8^2$, $16^2$, $32^2$ and $64^2$. We sum across attention heads following~\cite{tian2024diffuse, zhang2024deciphering}.
We aggregate over resolutions by upsampling to $64^2 \times 64^2$ followed by a proportional weighted sum, empirically defined in DiffSeg~\cite{tian2024diffuse}, producing an aggregated self-attention matrix 
%
    $P = \sum_{r} w_r \cdot S^{(r)}.$
%
Each matrix $S^{(r)}$ is row-normalized through softmax, and since the weights $w_r$ sum to one, the aggregated matrix P maintains row-normalization.

\subsection{Self-attention defines a random walk}
\label{subsec:method-attention-random-walk}

While traditional NCut requires constructing an adjacency matrix $A$, from which the random-walk interpretation derives a degree-normalized (i.e. row-stochastic) transition matrix $P = D^{-1}A$, we observe that the aggregated self-attention $P$ 
already satisfies the mathematical properties of a random walk transition matrix, albeit without being derived explicitly from some adjacency matrix $A$.

As described in the previous section, our aggregated self-attention matrix maintains row-normalization through the softmax operation in the attention mechanism and weighted aggregation. Consequently, each row in $P$ represents a probability distribution, with elements $P_{i,j}$ giving the probability of transitioning from patch $i$ to patch $j$ in a random walk across the image.

It is therefore possible to directly apply Normalized Cuts theory in the Random Walk reframing. We use the transition matrix as the input to recursive NCut as proposed by Shi and Malik~\cite{shi2000normalized}, as described in Section~\ref{subsec:method-ncut}, by adapting the algorithm to use the eigenvectors of the transition probabilities $P$ (c.f.~\cite{NIPS2000_069654d5}) instead of an adjacency matrix $A$. Since $P$ is not necessarily symmetric, to estimate the second eigenvector, we apply power iteration with one deflation.

This formulation provides a principled approach to leveraging self-attention directly for segmentation, without requiring the construction of an intermediate adjacency matrix or similarity metric, while still benefiting from the theoretical guarantees of spectral clustering.

\subsection{Adjacency matrices from feature similarity}
\label{subsec:method-feature-adjacency}

Beyond directly using self-attention as transition probabilities, we can construct traditional adjacency matrices by treating each patch's attention pattern as a feature embedding. While DiffSeg uses Kullback-Leibler divergence between attention distributions, we employ either dot product or cosine similarity (empirically contrasted in Section~\ref{subsec:exp-adjacency}), which can be computed more efficiently.

These constructions yield a positive-valued, symmetric adjacency matrix $A$ suitable for the conventional NCut formulation. This allows for efficient direct eigendecomposition rather than power iteration.

At the same time, we can exploit the random-walk property of the self-attention matrix established in Section~\ref{subsec:method-attention-random-walk} prior to constructing the adjacency matrix. By computing $P^k$ through matrix exponentiation, each entry $(P^k)_{ij}$ represents the probability of transitioning from patch $i$ to $j$ in exactly $k$ steps~\cite{newman2018networks}. This $k$-step transition probability aggregates all possible walks of length $k$ between patches $i$ and $j$, capturing information flow through intermediate patches.

The exponent $k$ serves as a scale parameter: smaller values preserve local attention patterns, while larger values enable broader information diffusion across the image, connecting semantically related but spatially distant patches (see first row of Figure~\ref{fig:qualitative-comparison-power-settings}). We evaluate several values of $k$ in Section~\ref{subsec:power-walks}, showing how this parameter effectively controls the hierarchical level of segmentation.

\begin{figure}[!h]
\begin{center}
    \resizebox{\linewidth}{!}{%
        \begin{tikzpicture}
          \node[inner sep=0] (image1) {\includegraphics[height=2.5cm]{figures/random_walk_baseline/baseball_input.png}};
        \end{tikzpicture}%
        \hspace{0.1em}
        \begin{tikzpicture}
          \node[inner sep=0] (image2) {\includegraphics[height=2.5cm]{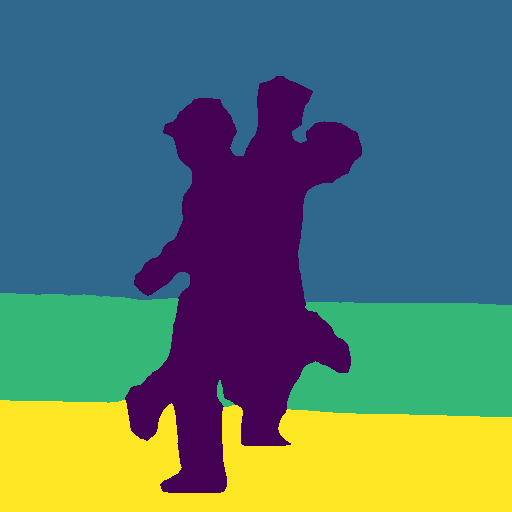}};
        \end{tikzpicture}%
        \begin{tikzpicture}
          \node[inner sep=0] (image3) {\includegraphics[height=2.5cm]{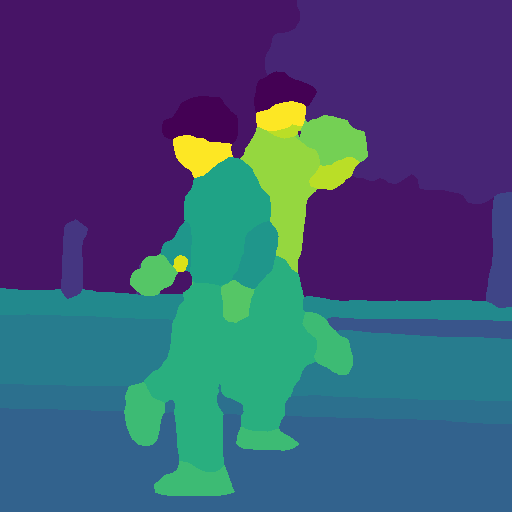}};
        \end{tikzpicture}%
        \begin{tikzpicture}
          \node[inner sep=0] (image4) {\includegraphics[height=2.5cm]{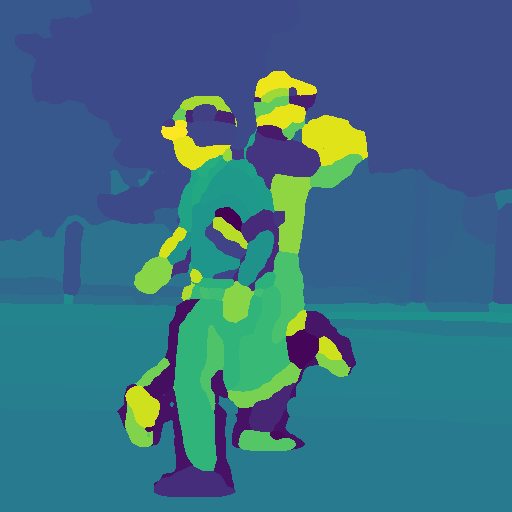}};
        \end{tikzpicture}%
        \hspace{0.1em}
        \begin{tikzpicture}
          \node[inner sep=0] (image5) {\includegraphics[height=2.5cm]{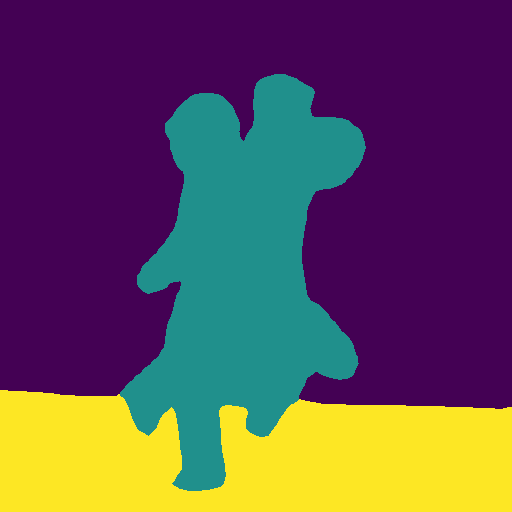}};
        \end{tikzpicture}%
        \begin{tikzpicture}
          \node[inner sep=0] (image6) {\includegraphics[height=2.5cm]{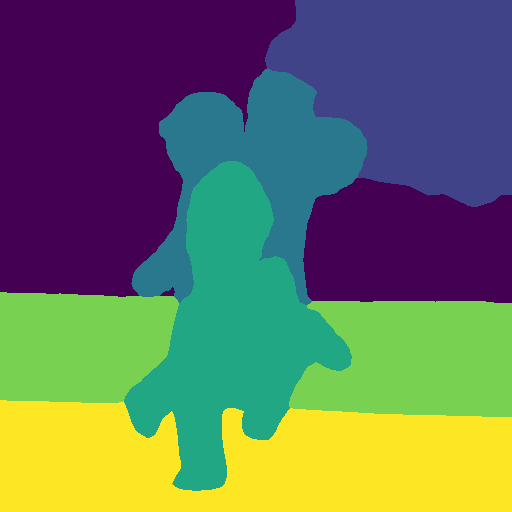}};
        \end{tikzpicture}%
        \begin{tikzpicture}
          \node[inner sep=0] (image7) {\includegraphics[height=2.5cm]{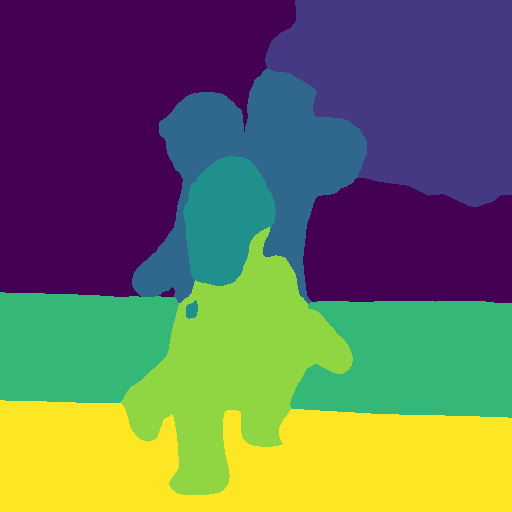}};
        \end{tikzpicture}%
        \hspace{0.1em}
        \begin{tikzpicture}
          \node[inner sep=0] (image8) {\includegraphics[height=2.5cm]{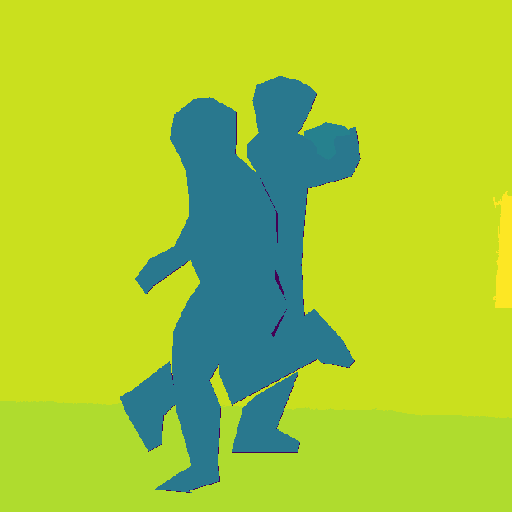}};
        \end{tikzpicture}%
    }
    \resizebox{\linewidth}{!}{%
        \begin{tikzpicture}
          \node[inner sep=0] (image1) {\includegraphics[height=2.5cm]{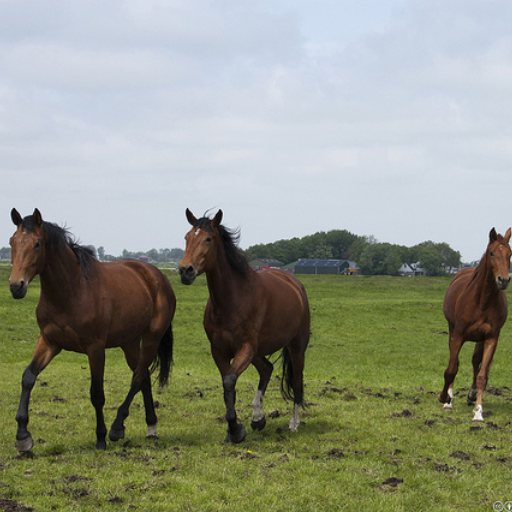}};
        \end{tikzpicture}%
        \hspace{0.1em}
        \begin{tikzpicture}
          \node[inner sep=0] (image2) {\includegraphics[height=2.5cm]{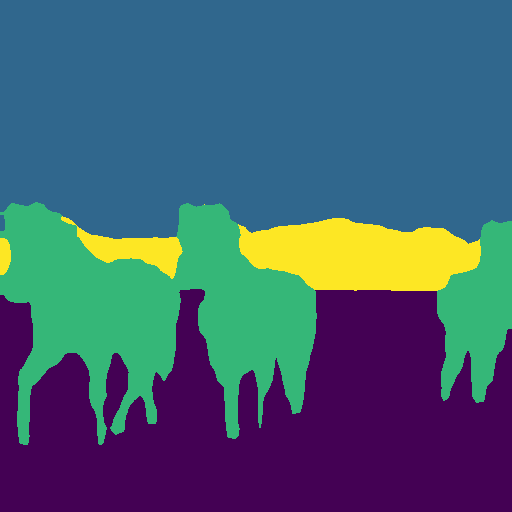}};
        \end{tikzpicture}%
        \begin{tikzpicture}
          \node[inner sep=0] (image3) {\includegraphics[height=2.5cm]{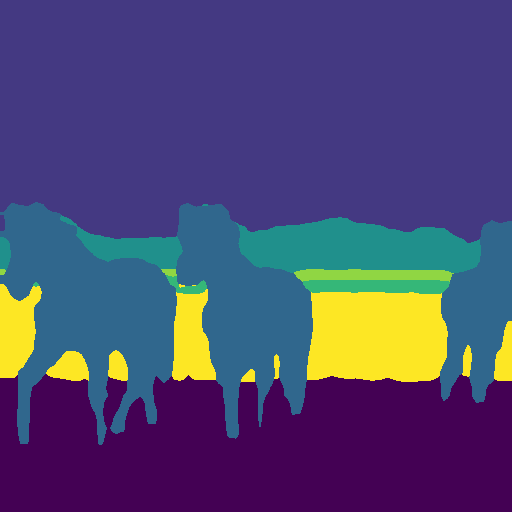}};
        \end{tikzpicture}%
        \begin{tikzpicture}
          \node[inner sep=0] (image4) {\includegraphics[height=2.5cm]{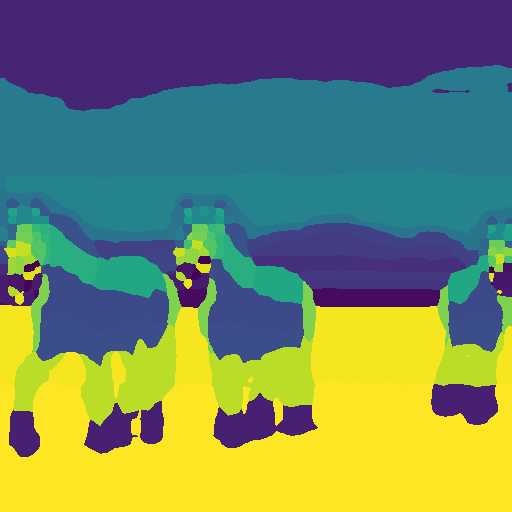}};
        \end{tikzpicture}%
        \hspace{0.1em}
        \begin{tikzpicture}
          \node[inner sep=0] (image5) {\includegraphics[height=2.5cm]{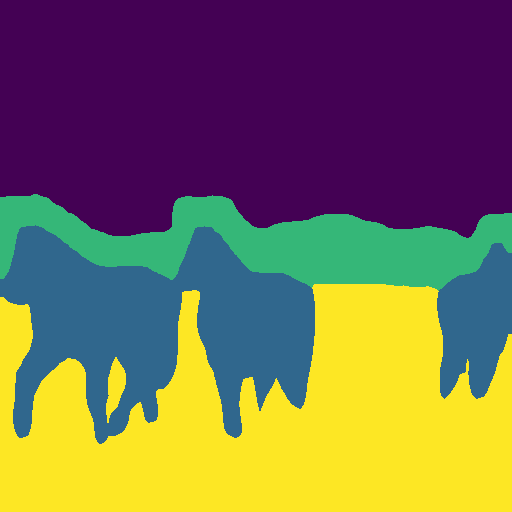}};
        \end{tikzpicture}%
        \begin{tikzpicture}
          \node[inner sep=0] (image6) {\includegraphics[height=2.5cm]{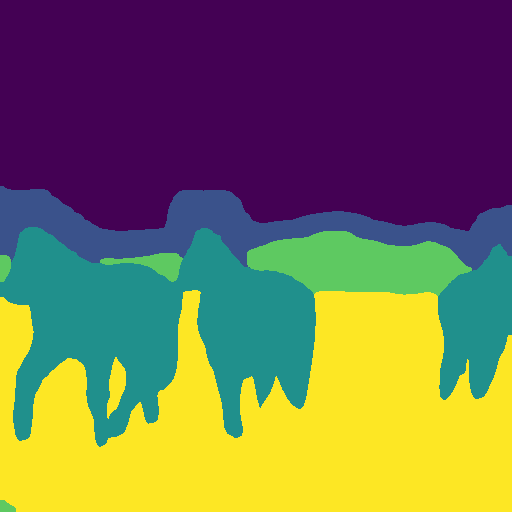}};
        \end{tikzpicture}%
        \begin{tikzpicture}
          \node[inner sep=0] (image7) {\includegraphics[height=2.5cm]{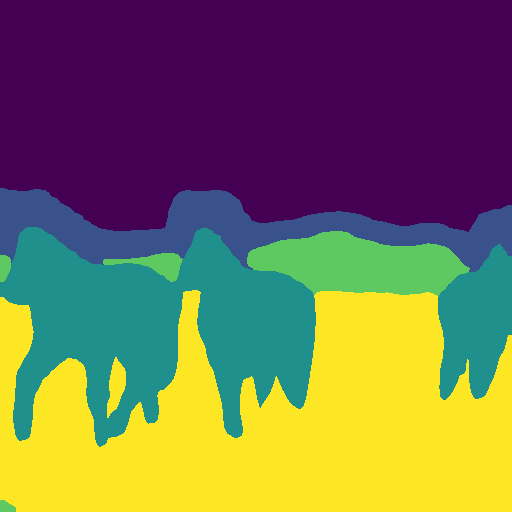}};
        \end{tikzpicture}%
        \hspace{0.1em}
        \begin{tikzpicture}
          \node[inner sep=0] (image8) {\includegraphics[height=2.5cm]{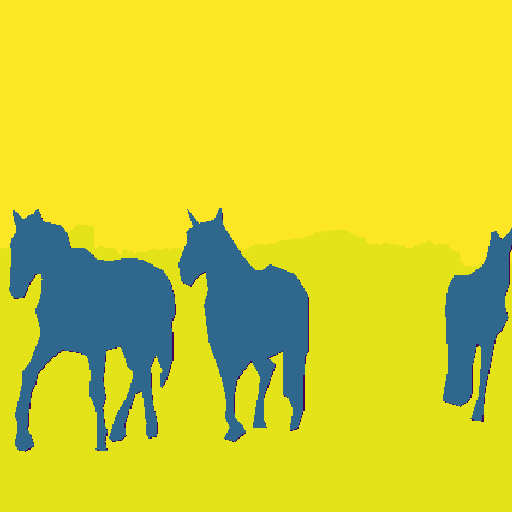}};
        \end{tikzpicture}%
    }    
   
    \resizebox{\linewidth}{!}{%
        \begin{tikzpicture}
          \node[inner sep=0] (image1) {\includegraphics[height=2.5cm]{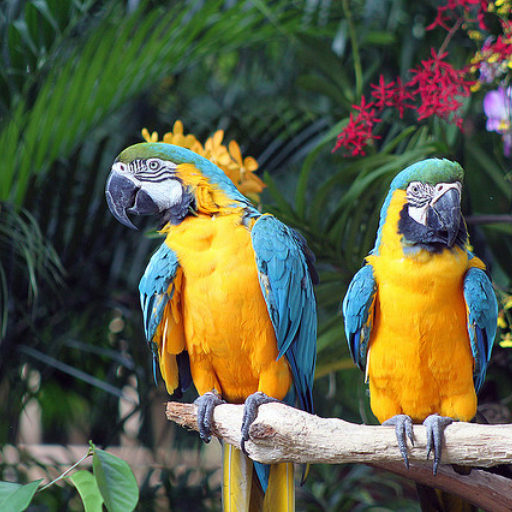}};
            \node[anchor=north] at ($(image1.south)+(0.0,0.0)$) {\textcolor{GHOST}{p}\Large{Input}\textcolor{GHOST}{p}};
        \end{tikzpicture}%
        \hspace{0.1em}
        \begin{tikzpicture}
          \node[inner sep=0] (image2) {\includegraphics[height=2.5cm]{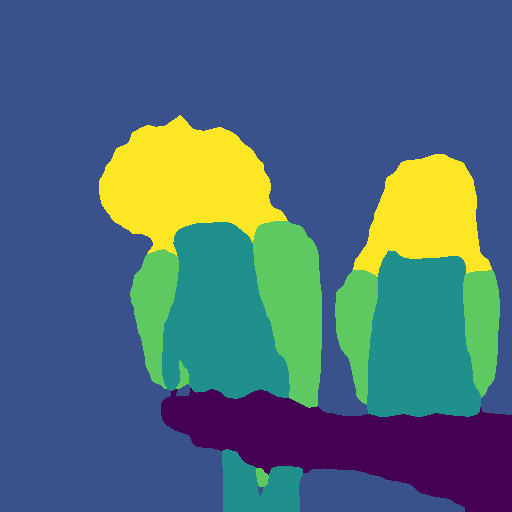}};
          \node[anchor=north] at ($(image2.south)+(0.0,0.0)$) {\textcolor{GHOST}{j}\Large{NCut=0.3}\textcolor{GHOST}{j}};
        \end{tikzpicture}%
        \begin{tikzpicture}
          \node[inner sep=0] (image3) {\includegraphics[height=2.5cm]{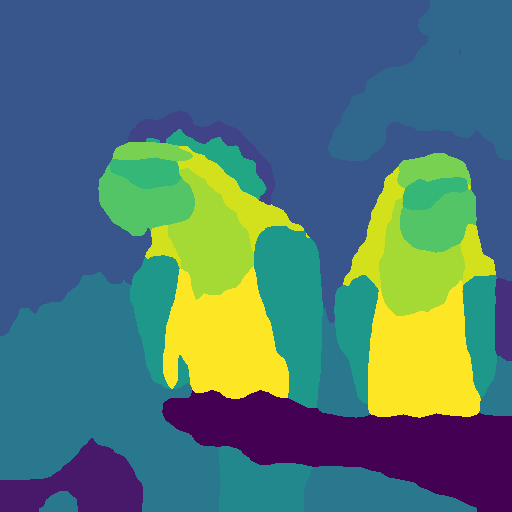}};
          \node[anchor=north] at ($(image3.south)+(0.0,0.0)$) {\textcolor{GHOST}{j}\Large{NCut=0.4}\textcolor{GHOST}{j}};
        \end{tikzpicture}%
        \begin{tikzpicture}
          \node[inner sep=0] (image4) {\includegraphics[height=2.5cm]{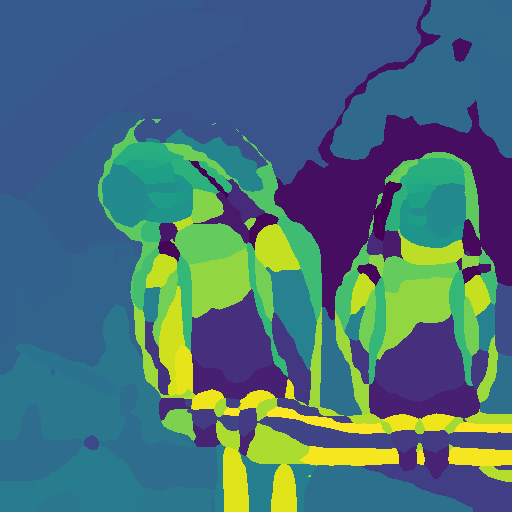}};
          \node[anchor=north] at ($(image4.south)+(0.0,0.0)$) {\textcolor{GHOST}{j}\Large{NCut=0.5}\textcolor{GHOST}{j}};
        \end{tikzpicture}%
        \hspace{0.1em}
        \begin{tikzpicture}
          \node[inner sep=0] (image5) {\includegraphics[height=2.5cm]{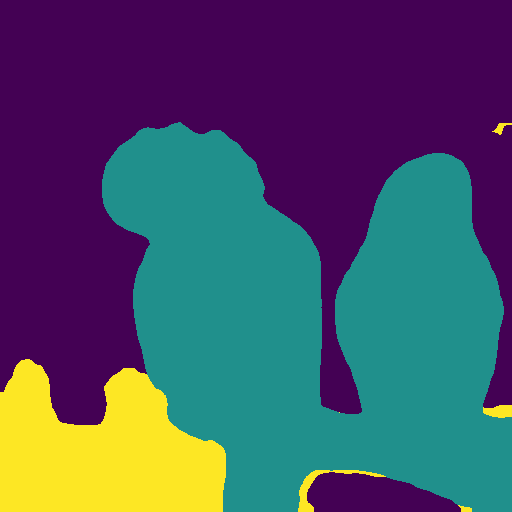}};
          \node[anchor=north] at ($(image5.south)+(0.0,0.0)$) {\textcolor{GHOST}{p}\Large{KL=1.1}\textcolor{GHOST}{p}};
        \end{tikzpicture}%
        \begin{tikzpicture}
          \node[inner sep=0] (image6) {\includegraphics[height=2.5cm]{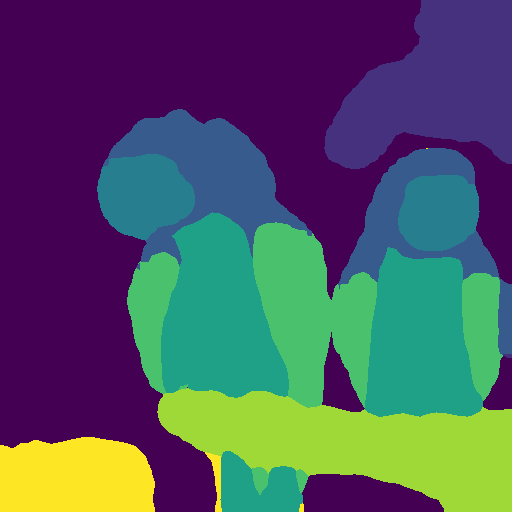}};
          \node[anchor=north] at ($(image6.south)+(0.0,0.0)$) {\textcolor{GHOST}{p}\Large{KL=0.9}\textcolor{GHOST}{p}};
        \end{tikzpicture}%
        \begin{tikzpicture}
          \node[inner sep=0] (image7) {\includegraphics[height=2.5cm]{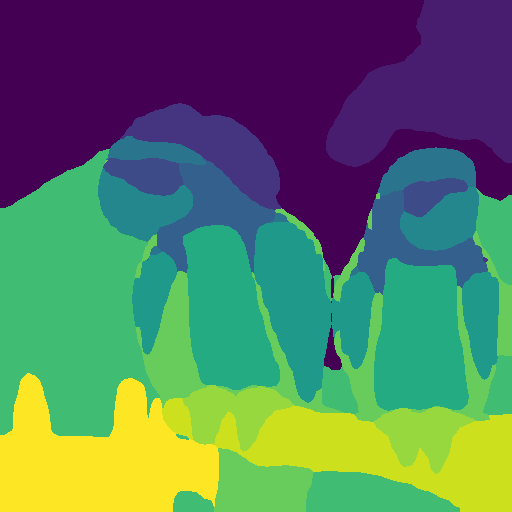}};
          \node[anchor=north] at ($(image7.south)+(0.0,0.0)$) {\textcolor{GHOST}{p}\Large{KL=0.7}\textcolor{GHOST}{p}};
        \end{tikzpicture}%
        \hspace{0.1em}
        \begin{tikzpicture}
          \node[inner sep=0] (image8) {\includegraphics[height=2.5cm]{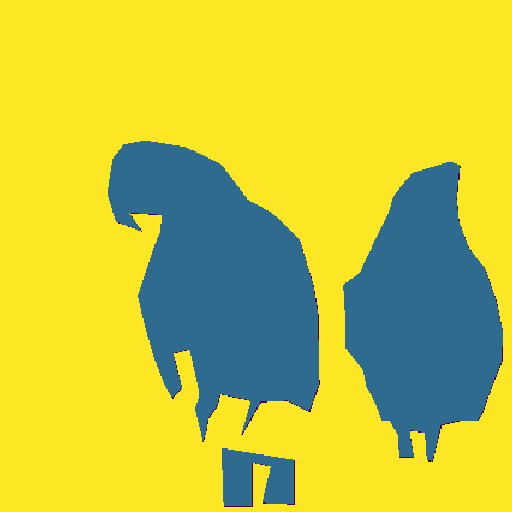}};
          \node[anchor=north] at ($(image8.south)+(0.0,0.0)$) {\textcolor{GHOST}{p}\Large{GT}\textcolor{GHOST}{p}};
        \end{tikzpicture}%
    }
\end{center}
\vspace{-1em}
\caption{Qualitative comparison on COCO-Stuff-27 between our simple Random Walk approach for different NCut cost thresholds (columns 2-4) and DiffSeg for different Kullback-Leibler Divergence thresholds (columns 5-7).}
\vspace{-1.5em}
\label{fig:qualitative-random-walk}
\end{figure}

\subsection{Applying normalised cuts (NCut)}
\label{subsec:method-ncut}
So far, we have introduced two approaches to quantify semantic correlations across the image by (1) modeling the aggregated self-attention matrix as a k-step random walk among patches via matrix exponentiation, and
(2) constructing an adjacency matrix from patch-level feature similarities. As discussed in Section~\ref{subsec:relwork-spectral-clustering}, both introduced formulations allow us to apply Shi and Malik’s recursive Normalised Cuts (NCut) criterion~\cite{shi2000normalized} to partition the image. Regardless of whether a random walk or adjacency matrix is used, the algorithm proceeds as follows. (1) Solve the generalised eigenvalue problem (equation~\ref{eq:L_eigen}). (2) Threshold the Fiedler vector across test values, selecting the bipartition with lowest NCut value. (3) If the NCut cost exceeds a predefined threshold, return the current partitions. (4) Otherwise, recursively apply the algorithm to each newly formed partition.

\subsection{Adapting MAN-C}
\label{subsec:method-automatic-thresh}
A key challenge in Normalized Cuts is determining when to stop recursive partitioning. Manual threshold selection is burdensome and image-dependent. To address this, we adapt the Minimum Average Node Cut (MAN-C) algorithm \cite{wang2007graph}, detailed in Section~\ref{sec:related_works}. MAN-C works well for sparse graphs, where there are few edges. However, our graphs derived from self-attention similarities have very high edge density, and in this case the MAN-C threshold reduces to $\frac{T}{n-1}$ i.e. it is scaled by the number of nodes---therefore, a minimum cut will almost always be larger, since it's not appropriately scaled w.r.t. the threshold anymore.

Hence, we propose and evaluate (Section~\ref{subsec:method-automatic-thresh}) two alternative adjustments: 
\begin{enumerate} 
\item Directly compare NCut cost to the MAN-C threshold;
\item Compare to a \emph{scaled} minimum cut, $\frac{\text{min-cut}(G)}{T}$, which scales the minimum cut relative to the graph’s total edge weight $T$, to place it in the same scale as the MAN-C threshold.
\end{enumerate}
Both approaches are designed to keep the adaptive nature of MAN-C while handling the dense nature of our similarity graphs.

\subsection{Segmentation aggregation and upsampling}
\label{subsec:method-mask-upsampling}

After recursive partitioning, we aggregate the NCut partitions into unique segments and calculate the average feature vector for each region, resulting in $K$ averaged PDFs for $K$ segments of shape $N \times N$, where $N$ is the latent size (number of latent patches). We upsample these segmentations to the original image resolution by first applying bilinear interpolation to the original features---$512 \times 512 \times N^2$, then computing cosine similarity between each upsampled pixel feature and each segment's average distribution. Each pixel is assigned to its closest segment, producing the final segmentation map.

\section{Experiments}
\label{sec:experiments}

\paragraph{Datasets.}
For our main experiments, we follow ~\cite{cho2021picie, ji2019invariant, tian2024diffuse, namekata2024emerdiff} and use a subset~\cite{ji2019invariant} of COCO-Stuff-27~\cite{caesar2018coco} validation data. It consists of 2174 images, combining 80 thing and 91 stuff categories from COCO-Stuff-171 into 27 classes. Our final approach is also evaluated on Cityscapes~\cite{Cordts2016Cityscapes}, following DiffSeg \cite{tian2024diffuse}. We additionally evaluate on ADE20K~\cite{zhou2017scene}, and use it for our comparison with DiffCut~\cite{couairon2024zero}.

\paragraph{Metrics.}
To quantitatively evaluate unsupervised segmentation, we report Mean Pixel Accuracy, Mean Intersection over Union (mIoU), as done in previous works~\cite{tian2024diffuse, namekata2024emerdiff, hamilton2022unsupervised}, in addition to F1 score.
Since class labels are not predicted, we use Hungarian matching to assign predicted segments to their closest Ground Truth segments, following standard practice~\cite{tian2024diffuse, namekata2024emerdiff, Sick_2024_CVPR}. 
This penalizes oversegmentation, since when there are more predicted segments than Ground Truth ones, the remaining predicted segments are considered false positives. 
We note that due to the varying granularity of human annotated segmentation labels, this standard evaluation strategy is far from perfect. For example, in Figure~\ref{fig:qualitative-random-walk}, row 1, the Ground Truth merges the human figures, along with attributes such as gloves and helmets into a single annotation. Our method can match this annotation granularity with a lower NCut threshold (row 1, column 2), or separate the baseball glove, along with the hats, shoes, trousers, shirts, legs and arms into their own segments when the threshold is increased. We dicuss how this affects quantitative evaluation in the Supplementary, Section A. 

\subsection{Random Walk Baseline}


As outlined in Section~\ref{subsec:method-attention-random-walk}, our simple baseline approach uses the aggregated self-attention matrices directly as a random walk transition matrix and applies recursive Normalized Cuts over it. We show examples from several thresholds over the NCut cost as a stopping criterion in Figure~\ref{fig:qualitative-random-walk}, where we also compare it to results from DiffSeg for their best-perfoming KLD threshold ($1.1$, column 5) along with two lower values. Since Normalised Cuts is inherently a top-down segmentation approach, we note how a higher NCut threshold increases granularity in the predicted segments, which generally follow a hierarchical structure; for DiffSeg, lower KLD threshold has a similar effect, as the approach merges segments in a bottom-up fashion. Overall, our approach's predictions are more precise, and demonstrate a clear hierarchical semantic relationship between segments across thresholds, even in the case of extreme oversegmention---for instance, the horses legs, hooves, backs and faces are all grouped together in row 2, column 4. DiffSeg predictions, on the other hand, suffer from pronounced horizontal banding artefacts (row 2, 5).
Quantitative results for NCut threshold set to $0.3$ in Table~\ref{tab:quantitative-sota} show that this simple approach already performs almost as well as the state-of-the-art DiffSeg, particularly with respect to mIoU and accuracy.

\subsection{Self-Attention adjacency}
\label{subsec:exp-adjacency}

\begin{figure}[!t]
    \centering
    \begin{subfigure}[b]{0.32\linewidth}
        \includegraphics[width=\linewidth]{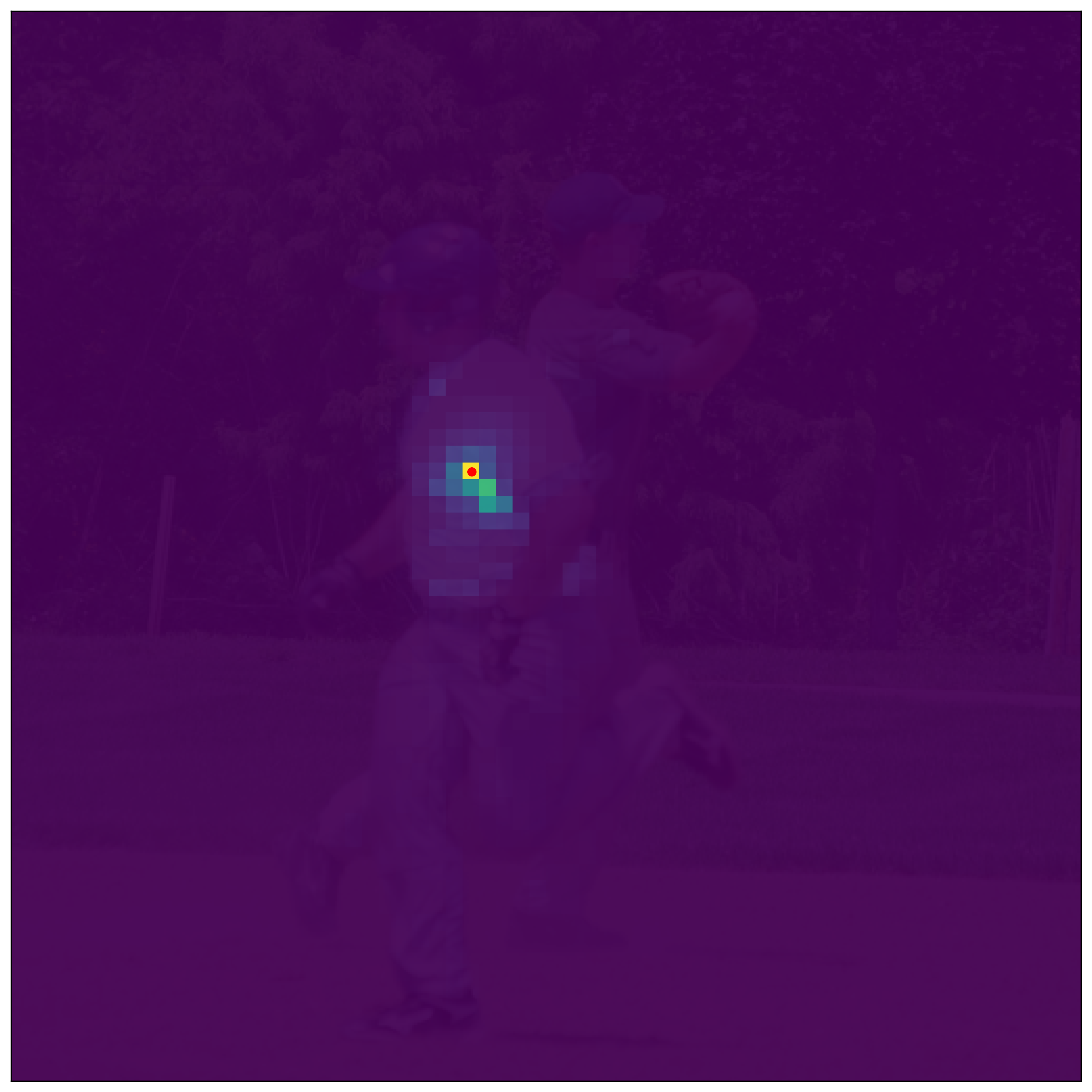}
        \caption{Self-attention PDF (aggregated)}
    \end{subfigure}
    \hfill
    \begin{subfigure}[b]{0.32\linewidth}
        \includegraphics[width=\linewidth]{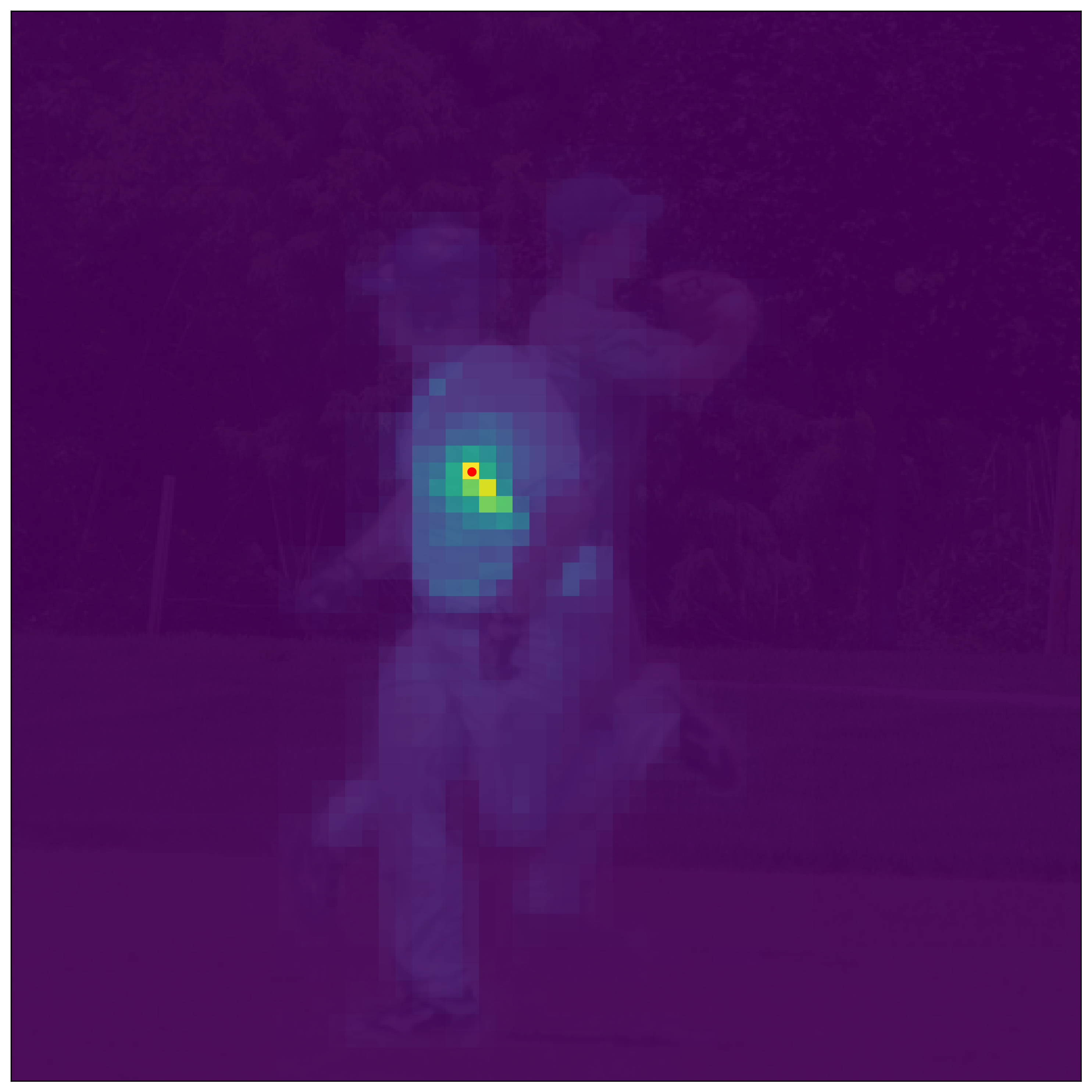}
        \caption{Dot product adjacency}
    \end{subfigure}
    \hfill
    \begin{subfigure}[b]{0.32\linewidth}
        \includegraphics[width=\linewidth]{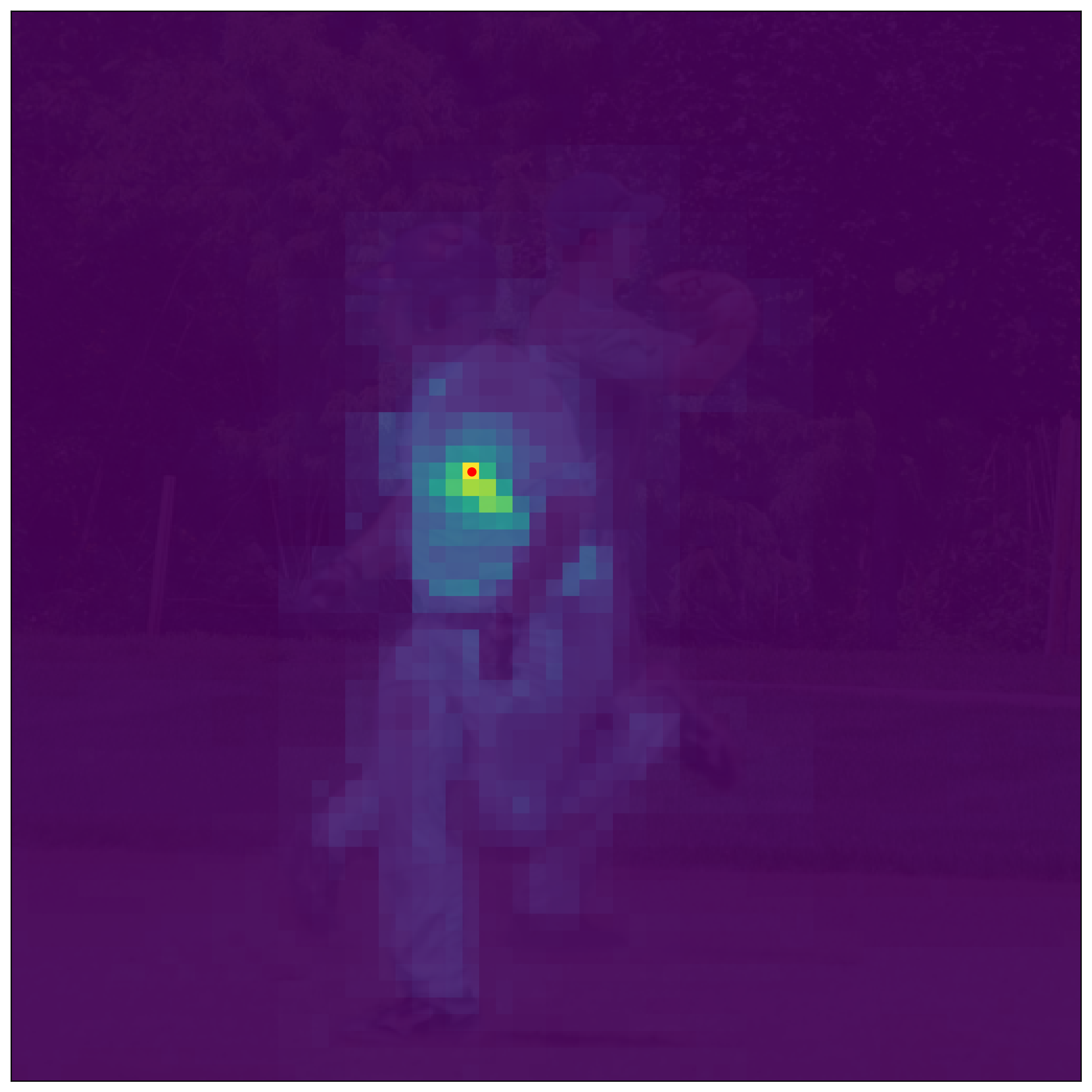}
        \caption{Cosine similarity adjacency}
    \end{subfigure}
    \caption{Self-attention PDF (aggregated) (a) and corresponding dot product (b) and cosine similarity (c) adjacency values for a single patch.}
    \label{fig:spatial_views}
    \vspace{-1em}
\end{figure}

The previous experiment motivates a natural follow-up: how can we best construct a traditional NCuts adjacency matrix based on the similarity between self-attention activations? Here, we validate the two variants proposed in Section~\ref{subsec:method-feature-adjacency}: constructing $A$ as the dot-product between the self-attention PDFs, or as their cosine similarity (normalised dot product). Both variants result in a positive-valued symmetric adjacency matrix $A$, since self-attention is already probability-normalised. We compare the results of applying Normalised Cuts over each variant across three different NCut Thresholds in Table~\ref{tab:adjacency-dot-vs-cos}, where the Dot Product variant is the clear leader---note that when the NCut threshold is set to 0.4, it already outperforms state-of-the-art DiffSeg in terms of both mIoU and Accuracy.

\begin{figure}[!h]
\begin{center}
    \resizebox{\linewidth}{!}{%
        \begin{tikzpicture}
          \node[inner sep=0] (image1) {\includegraphics[height=2.5cm]{figures/random_walk_baseline/baseball_input.png}};
        \end{tikzpicture}%
        \hspace{0.1em}
        \begin{tikzpicture}
          \node[inner sep=0] (image2) {\includegraphics[height=2.5cm]{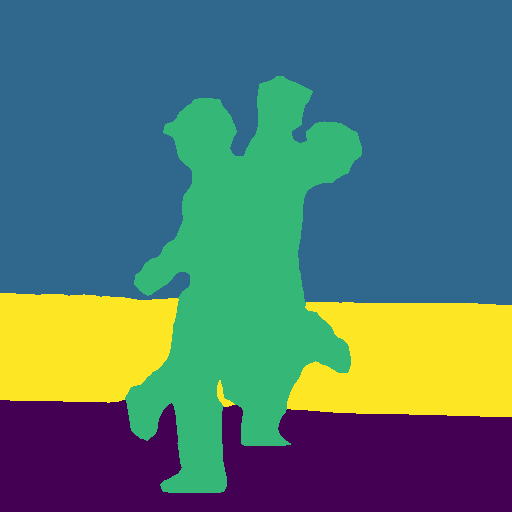}};
        \end{tikzpicture}%
        \begin{tikzpicture}
          \node[inner sep=0] (image3) {\includegraphics[height=2.5cm]{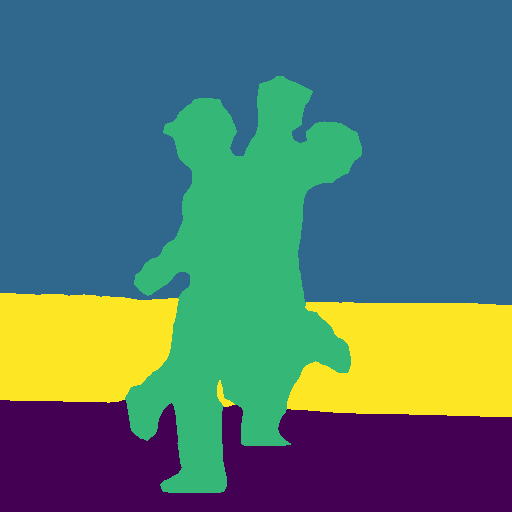}};
        \end{tikzpicture}%
        \begin{tikzpicture}
          \node[inner sep=0] (image4) {\includegraphics[height=2.5cm]{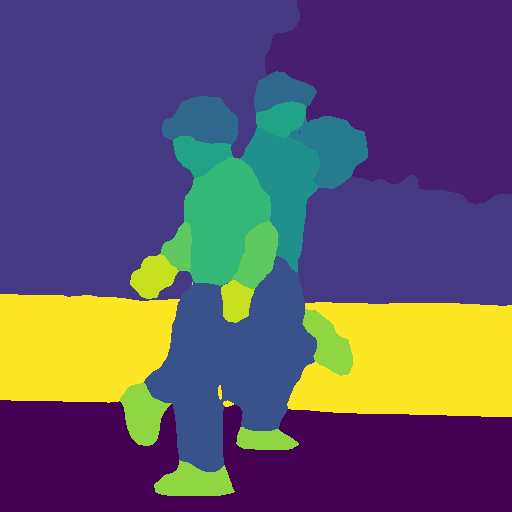}};
        \end{tikzpicture}%
        \hspace{0.1em}
        \begin{tikzpicture}
          \node[inner sep=0] (image5) {\includegraphics[height=2.5cm]{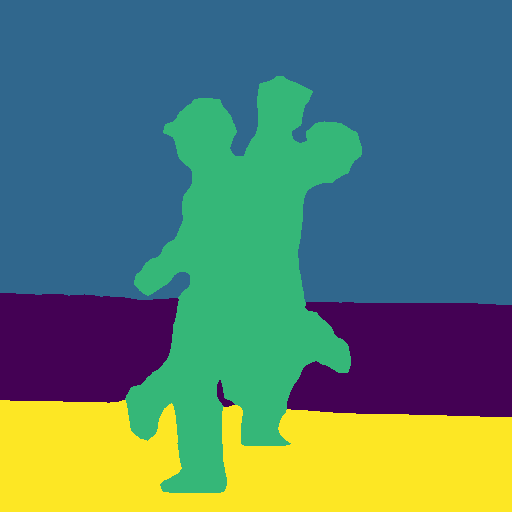}};
        \end{tikzpicture}%
        \begin{tikzpicture}
          \node[inner sep=0] (image6) {\includegraphics[height=2.5cm]{figures/adjacency-sim/baseball_dot5.png}};
        \end{tikzpicture}%
        \begin{tikzpicture}
          \node[inner sep=0] (image7) {\includegraphics[height=2.5cm]{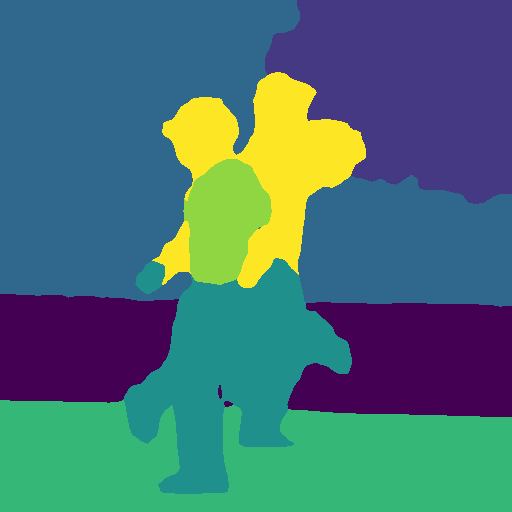}};
        \end{tikzpicture}%
        \hspace{0.1em}
        \begin{tikzpicture}
          \node[inner sep=0] (image8) {\includegraphics[height=2.5cm]{figures/random_walk_baseline/baseball_gt.png}};
        \end{tikzpicture}%
    }
    \resizebox{\linewidth}{!}{%
        \begin{tikzpicture}
          \node[inner sep=0] (image1) {\includegraphics[height=2.5cm]{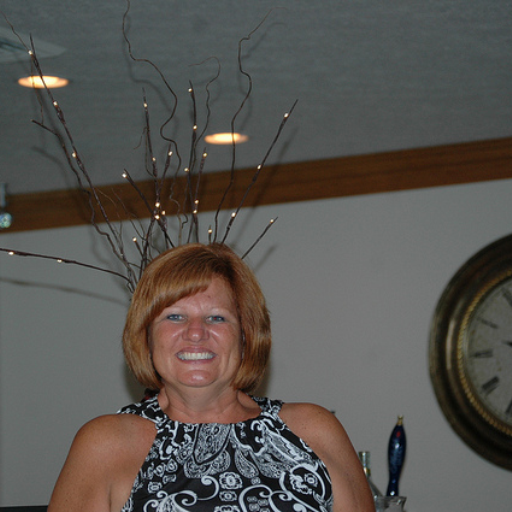}};
          \node[anchor=north] at ($(image1.south)+(0.0,0.0)$) {\textcolor{GHOST}{j}\Large{Input}\textcolor{GHOST}{j}};
        \end{tikzpicture}%
        \hspace{0.1em}
        \begin{tikzpicture}
          \node[inner sep=0] (image2) {\includegraphics[height=2.5cm]{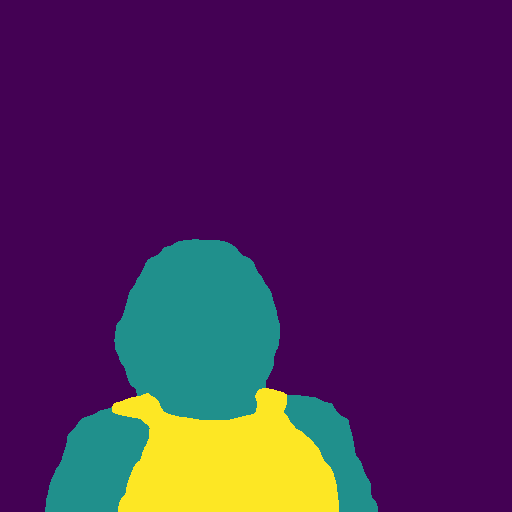}};
          \node[anchor=north] at ($(image2.south)+(0.0,0.0)$) {\textcolor{GHOST}{j}\Large{NCut=0.4}\textcolor{GHOST}{j}};
        \end{tikzpicture}%
        \begin{tikzpicture}
          \node[inner sep=0] (image3) {\includegraphics[height=2.5cm]{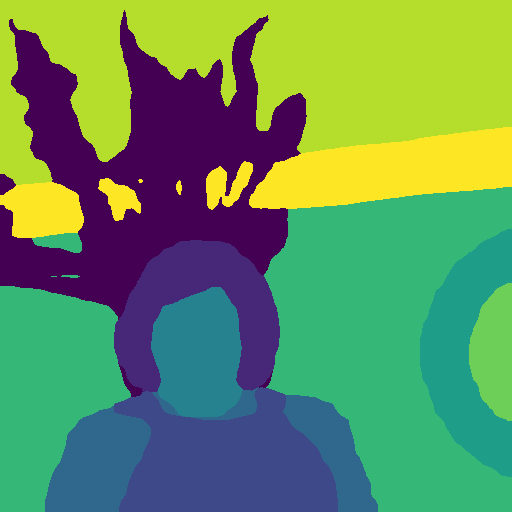}};
          \node[anchor=north] at ($(image3.south)+(0.0,0.0)$) {\textcolor{GHOST}{j}\Large{NCut=0.5}\textcolor{GHOST}{j}};
        \end{tikzpicture}%
        \begin{tikzpicture}
          \node[inner sep=0] (image4) {\includegraphics[height=2.5cm]{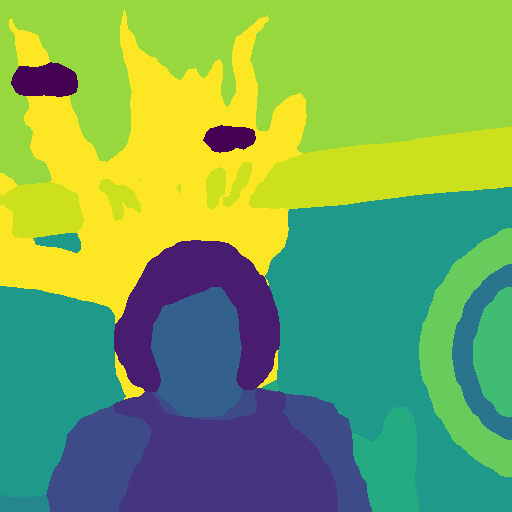}};
          \node[anchor=north] at ($(image4.south)+(0.0,0.0)$) {\textcolor{GHOST}{j}\Large{NCut=0.6}\textcolor{GHOST}{j}};
        \end{tikzpicture}%
        \hspace{0.1em}
        \begin{tikzpicture}
          \node[inner sep=0] (image5) {\includegraphics[height=2.5cm]{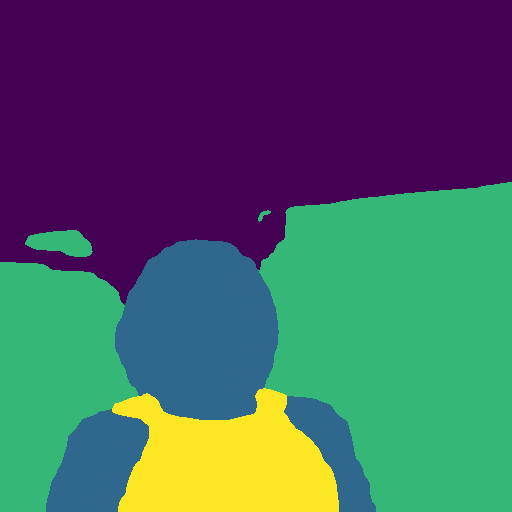}};
          \node[anchor=north] at ($(image4.south)+(0.0,0.0)$) {\textcolor{GHOST}{j}\Large{NCut=0.4}\textcolor{GHOST}{j}};
        \end{tikzpicture}%
        \begin{tikzpicture}
          \node[inner sep=0] (image6) {\includegraphics[height=2.5cm]{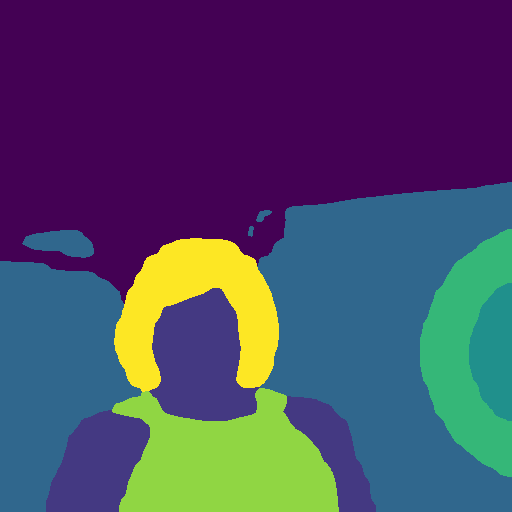}};
        \node[anchor=north] at ($(image4.south)+(0.0,0.0)$) {\textcolor{GHOST}{j}\Large{NCut=0.5}\textcolor{GHOST}{j}};
        \end{tikzpicture}%
        \begin{tikzpicture}
          \node[inner sep=0] (image7) {\includegraphics[height=2.5cm]{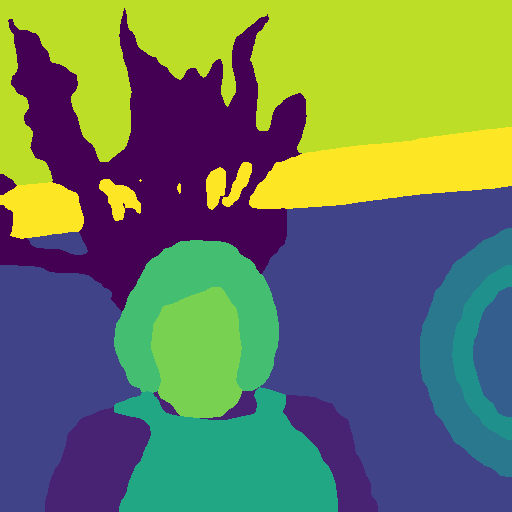}};
          \node[anchor=north] at ($(image4.south)+(0.0,0.0)$) {\textcolor{GHOST}{j}\Large{NCut=0.6}\textcolor{GHOST}{j}};
        \end{tikzpicture}%
        \hspace{0.1em}
        \begin{tikzpicture}
          \node[inner sep=0] (image8) {\includegraphics[height=2.5cm]{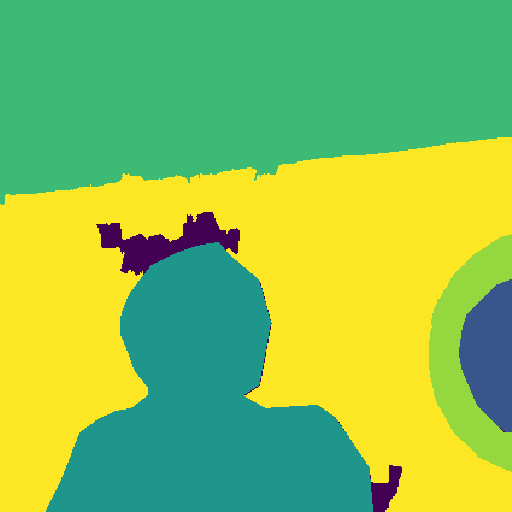}};
          \node[anchor=north] at ($(image5.south)+(0.0,0.0)$) {\textcolor{GHOST}{j}\Large{GT}\textcolor{GHOST}{j}};
        \end{tikzpicture}%
    }
\end{center}
\vspace{-12pt}
\caption{Qualitative comparison on COCO-Stuff-27 between NCut over a dot-product Adjacency matrix (columns 2-4) and over a cosine similarity Adjacency matrix (columns 5-7) across a range of NCut thresholds.}
\label{fig:qualitative-adjacency-similarity}
\end{figure}

Examining the spatial distribution plots of self-attention compared to the two adjacency variants for a given patch marked by red dot in Figure~\ref{fig:spatial_views}, we can see the intuition behind why dot product performs better. For a given query position $q$, the self-attention produces a PDF $p_q$ that is sharply peaked at $q$ and its immediate neighborhood $N(q)$, with near-zero values elsewhere, as shown in (a). The dot product $p_q^T p_r$ between two such distributions preserves this sparsity structure, showing high similarity only when neighborhoods overlap in (b). In contrast, cosine similarity $\frac{p_q^T p_r}{||p_q|| ||p_r||}$ normalizes away this locality-preserving property, artificially inflating similarities between distributions with disjoint support, as evidenced by more distributed similarities in (c), which form a halo around the object of interest. Hence, more uniformly distributed adjacency values would have a negative effect on Normalised Cuts, as it relies on having strong intra-segment connections and weak inter-segment connections; cosine similarity would therefore make the distinction between segments less pronounced and lead to arbitrary cuts, as seen in Figure~\ref{fig:qualitative-adjacency-similarity}.

\subsection{Hyperparameter-free NCut}
\label{subsec:automatic-ncuts}
In Section~\ref{subsec:method-automatic-thresh} we proposed a way to perform Normalised cuts without setting an NCut cost as a stopping criterion, making the approach completely hyperparameter-free.

\begin{table}[t]
\centering
\small
\begin{tabular}{@{}llccc@{}}
\toprule
\textbf{Model Variant} & \textbf{NCut Threshold} & \textbf{Acc $\uparrow$} & \textbf{F1 $\uparrow$} & \textbf{mIoU $\uparrow$} \\ \midrule
\multirow{3}{*}{\textbf{Dot Product}} & 0.4 & 74.0 & 58.4 & 44.0 \\
 & 0.5 & 71.6 & 56.7 & 43.1 \\
 & 0.6 & 60.0 & 45.2 & 32.1 \\ \midrule
\multirow{3}{*}{\textbf{Cosine Similarity}} & 0.4 & 72.4 & 56.0 & 41.7 \\
 & 0.5 & 71.5 & 56.0 & 42.2 \\
 & 0.6 & 62.9 & 47.7 & 34.4 \\ \bottomrule
\end{tabular}%
\caption{Quantitative results on COCO-Stuff-27 between Normalised Cuts performed on adjacency matrices build as dot product vs cosine distance between patch self-attention across three different NCut thresholds.}
\label{tab:adjacency-dot-vs-cos}
\vspace{-20pt}
\end{table}

\begin{figure*}[!t]
\begin{center} 
    \resizebox{\linewidth}{!}{%
        \begin{tikzpicture}
          \node[inner sep=0] (image1) {\includegraphics[height=2.5cm]{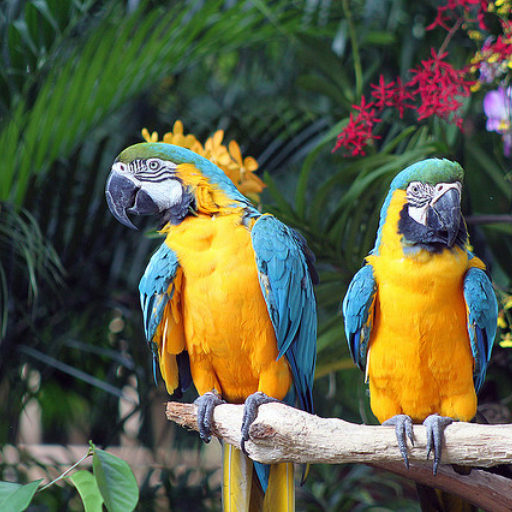}};
          \node[anchor=north] at ($(image1.south)+(0.0,0.0)$) {Input};
        \end{tikzpicture}%
        \hspace{0.1em}
        \begin{tikzpicture}
          \node[inner sep=0] (image2) {\includegraphics[height=2.5cm]{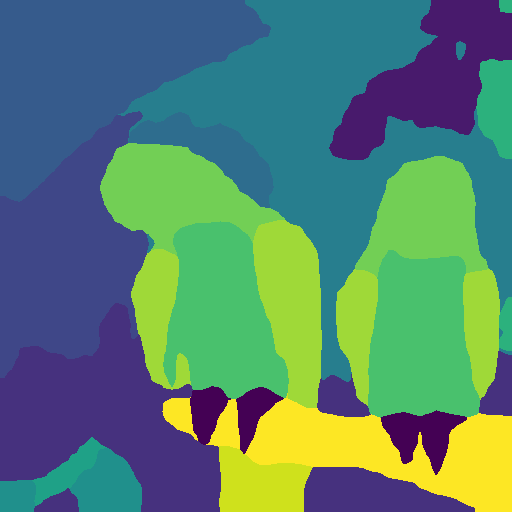}};
          \node[anchor=north] at ($(image2.south)+(0.0,0.0)$) {64, 32, 16, 8};
        \end{tikzpicture}%
        \begin{tikzpicture}
          \node[inner sep=0] (image3) {\includegraphics[height=2.5cm]{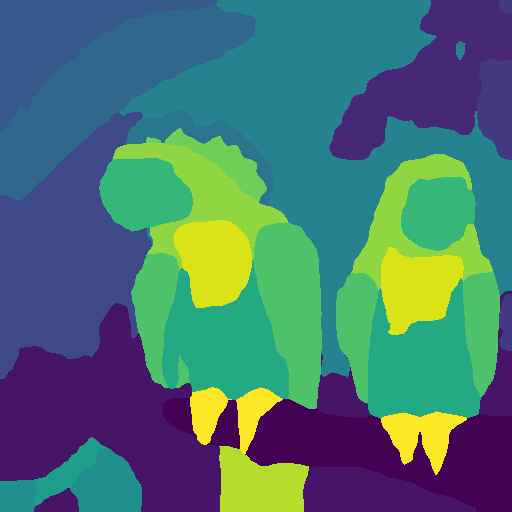}};
          \node[anchor=north] at ($(image3.south)+(0.0,0.0)$) {64, 32, 16};
        \end{tikzpicture}%
        \begin{tikzpicture}
          \node[inner sep=0] (image4) {\includegraphics[height=2.5cm]{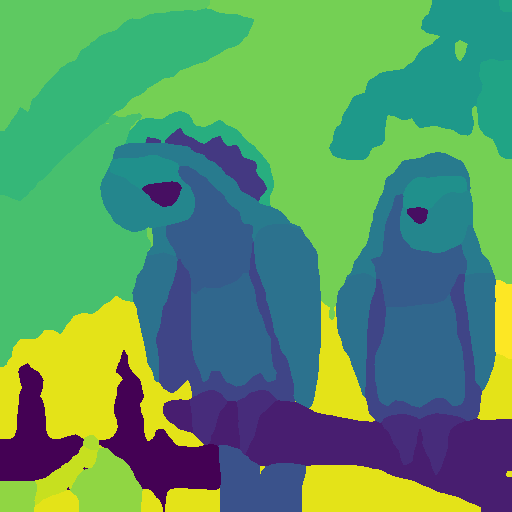}};
          \node[anchor=north] at ($(image4.south)+(0.0,0.0)$) {64, 32};
        \end{tikzpicture}%
        \begin{tikzpicture}
          \node[inner sep=0] (image5) {\includegraphics[height=2.5cm]{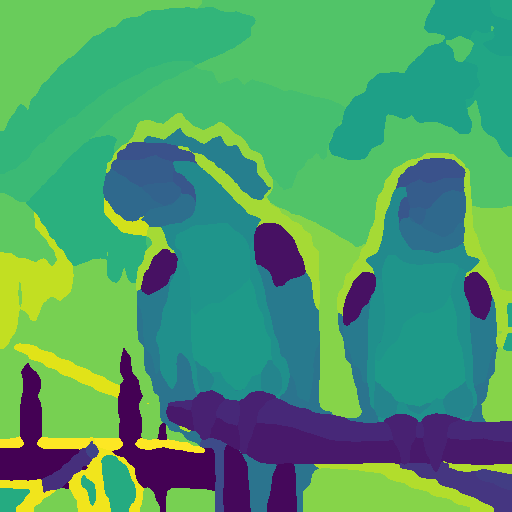}};
          \node[anchor=north] at ($(image5.south)+(0.0,0.0)$) {\textcolor{GHOST}{,}64\textcolor{GHOST}{,}};
        \end{tikzpicture}%
        \hspace{0.1em}
        \begin{tikzpicture}
          \node[inner sep=0] (image6) {\includegraphics[height=2.5cm]{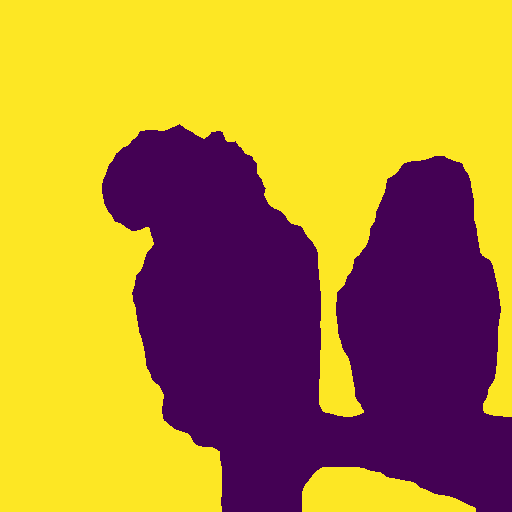}};
          \node[anchor=north] at ($(image6.south)+(0.0,0.0)$) {64, 32, 16, 8};
        \end{tikzpicture}%
        \begin{tikzpicture}
          \node[inner sep=0] (image7) {\includegraphics[height=2.5cm]{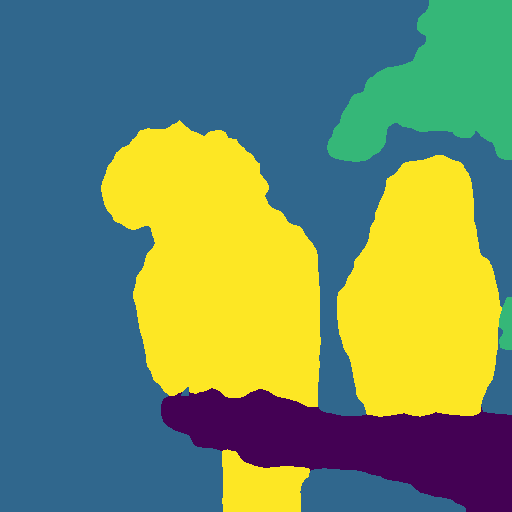}};
          \node[anchor=north] at ($(image7.south)+(0.0,0.0)$) {64, 32, 16};
        \end{tikzpicture}%
        \begin{tikzpicture}
          \node[inner sep=0] (image8) {\includegraphics[height=2.5cm]{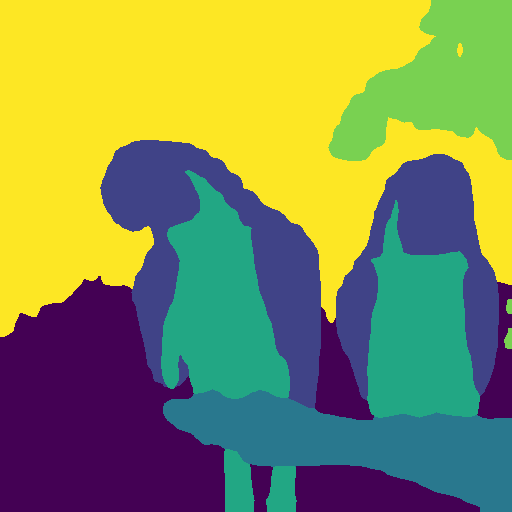}};
          \node[anchor=north] at ($(image8.south)+(0.0,0.0)$) {64, 32};
        \end{tikzpicture}%
        \begin{tikzpicture}
          \node[inner sep=0] (image9) {\includegraphics[height=2.5cm]{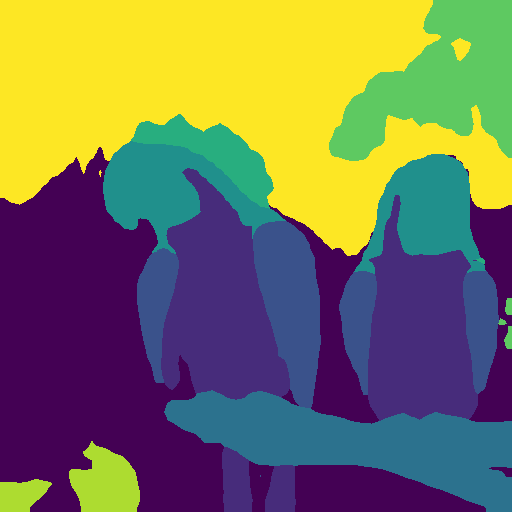}};
          \node[anchor=north] at ($(image9.south)+(0.0,0.0)$) {\textcolor{GHOST}{,}64\textcolor{GHOST}{,}};
        \end{tikzpicture}%
        \hspace{0.1em}
        \begin{tikzpicture}
          \node[inner sep=0] (image10) {\includegraphics[height=2.5cm]{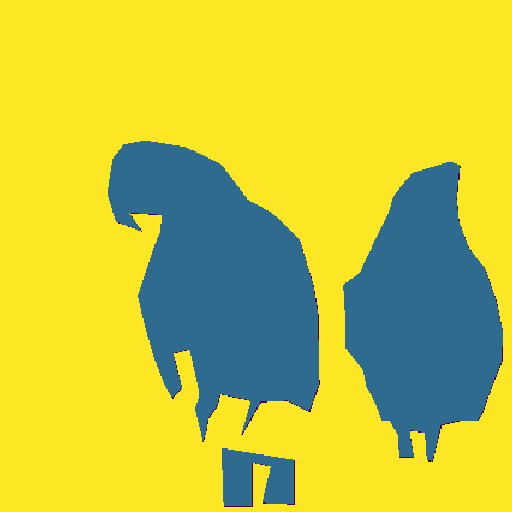}};
          \node[anchor=north] at ($(image10.south)+(0.0,0.0)$) {\textcolor{GHOST}{,}Ground Truth\textcolor{GHOST}{,}};
        \end{tikzpicture}%
    }
\end{center}
\vspace{-15pt}
\caption{Qualitative comparison of segmentation results for our two automatic thresholding approaches: Scaled MinCut (columns 2-5) and NCut (columns 6-9) across different self-attention resolution levels.}
\label{fig:qualitative-comparison-auto-thresh-resolutions}
\end{figure*}

Specifically, we suggested two versions of the MAN-C algorithm with thresholds adapted to the specific graph structure that arises from our adjacency matrices. We compare these two automatic thresholding approaches to assess their effectiveness in partitioning the self-attention adjacency graphs, using the dot product construction from the previous experiments. While the quantitative results in Table~\ref{tab:man-c-comparison} demonstrate that comparing the MAN-C criterion with NCuts is much more effective, we can see in the qualitative examples in Figure~\ref{fig:qualitative-comparison-auto-thresh-resolutions} that the segmentations proposed by thresholding with the scaled MinCut are not necessarily bad, as semantically meaningful groupings are clearly emerging; rather, they are too granular, and punished by the Hungarian matching due to the coarse annotations, as discussed earlier. Additionally, since the approach is now free of hyperparameters, we follow up with an ablation study on the self-attention resolution levels ${8, 16, 32, 64}$ by progressively removing each, starting from the lowest resolution. This allows us to evaluate how different self-attention scales contribute to segmentation quality. Removing lower resolutions increases the granularity of predictions for both methods, as seen in Figure~\ref{fig:qualitative-comparison-auto-thresh-resolutions}, and decreases the metrics they achieve; yet the qualitative examples demonstrate preserved semantic coherence even for the most granular predictions, such as those made with the Scaled MinCut threshold. Overall, our quantitative results show that we outperform DiffSeg, which optimises several dataset-specific hyperparameters (merging level, number of anchors, and KLD merging threshold) in a completely hyperparameter-free setting. 

\begin{table}[!t]
\small
\centering
\resizebox{\columnwidth}{!}{
\begin{tabular*}{\columnwidth}{@{\extracolsep{\fill}}cccccc@{}}
\toprule
\textbf{steps} k & 1 & 2 & 3 & 4 & 5 \\ 
\midrule
\textbf{Acc $\uparrow$}  & 45.9 & 67.0 & 71.2 & 70.7 & 70.0 \\
\textbf{F1 $\uparrow$}   & 36.1 & 54.4 & 56.9 & 55.8 & 54.2 \\
\textbf{mIoU $\uparrow$} & 23.8 & 40.7 & 42.4 & 41.1 & 39.6 \\
\bottomrule
\end{tabular*}
}
\caption{Comparison for a range exponents (random walk steps k) applied to the self-attention transition matrix $P$ before constructing the adjacency matrix $A$. We employ Scaled MinCut for automatic threshold selection, considering all available resolutions.}
\label{tab:power-walks}
\vspace{-2em}
\end{table}

\subsection{Power Walks}
\label{subsec:power-walks}
\begin{figure}[!htpb]
\begin{center}
    \resizebox{\linewidth}{!}{%
        \begin{tikzpicture}
          \node[inner sep=0] (image1) {\includegraphics[height=2.5cm]{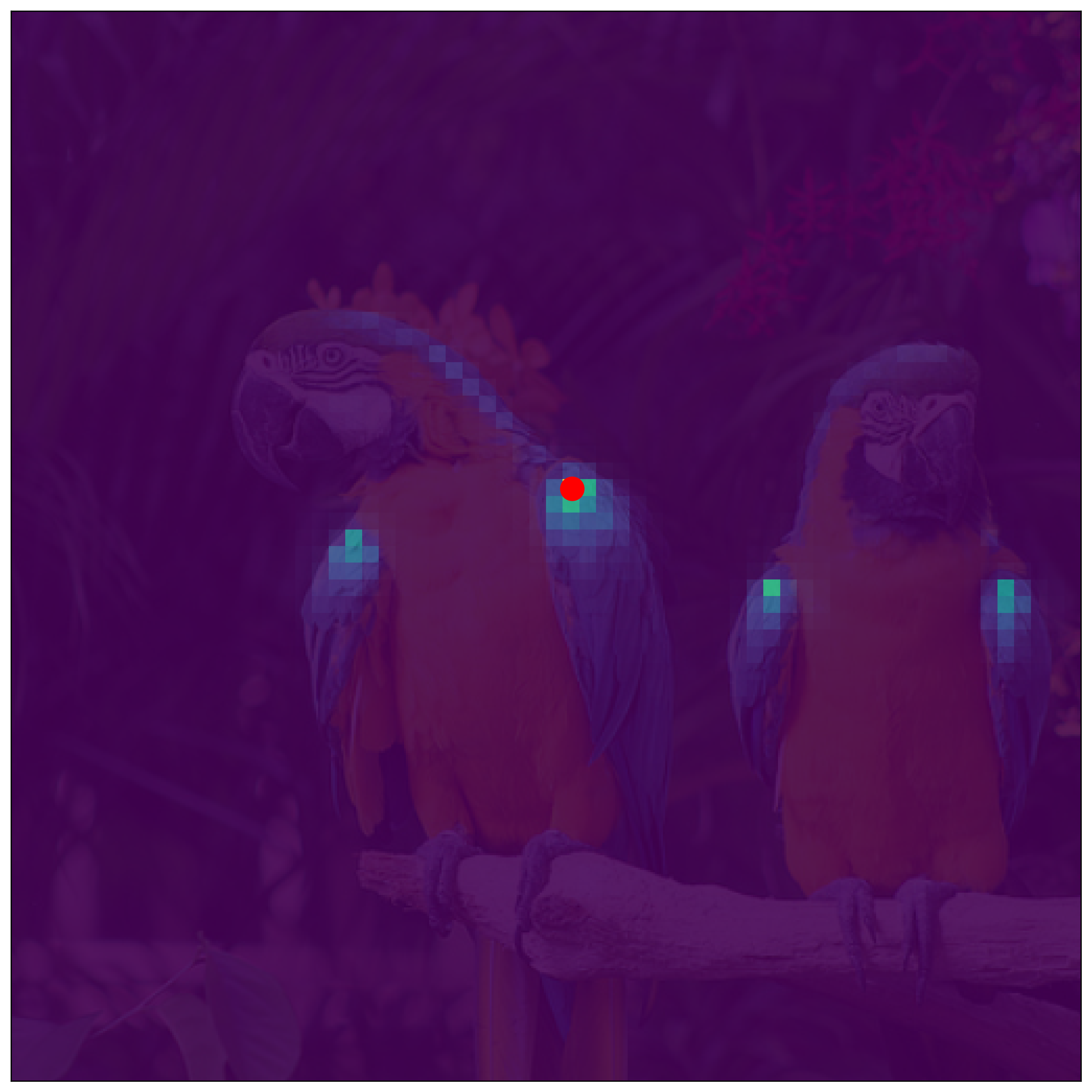}};
        \end{tikzpicture}%
        \hspace{0.1em}
        \begin{tikzpicture}
          \node[inner sep=0] (image2) {\includegraphics[height=2.5cm]{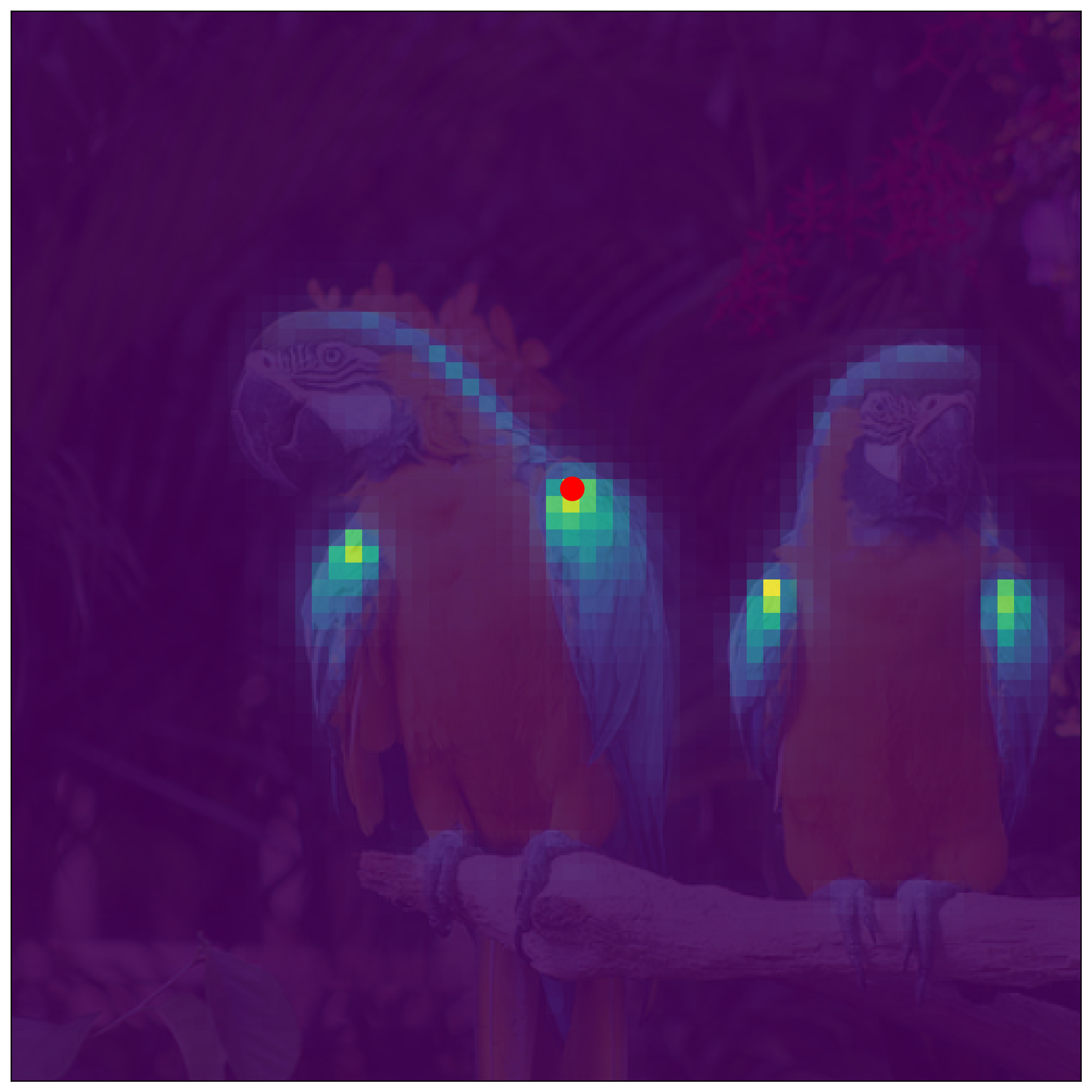}};
        \end{tikzpicture}%
        \begin{tikzpicture}
          \node[inner sep=0] (image3) {\includegraphics[height=2.5cm]{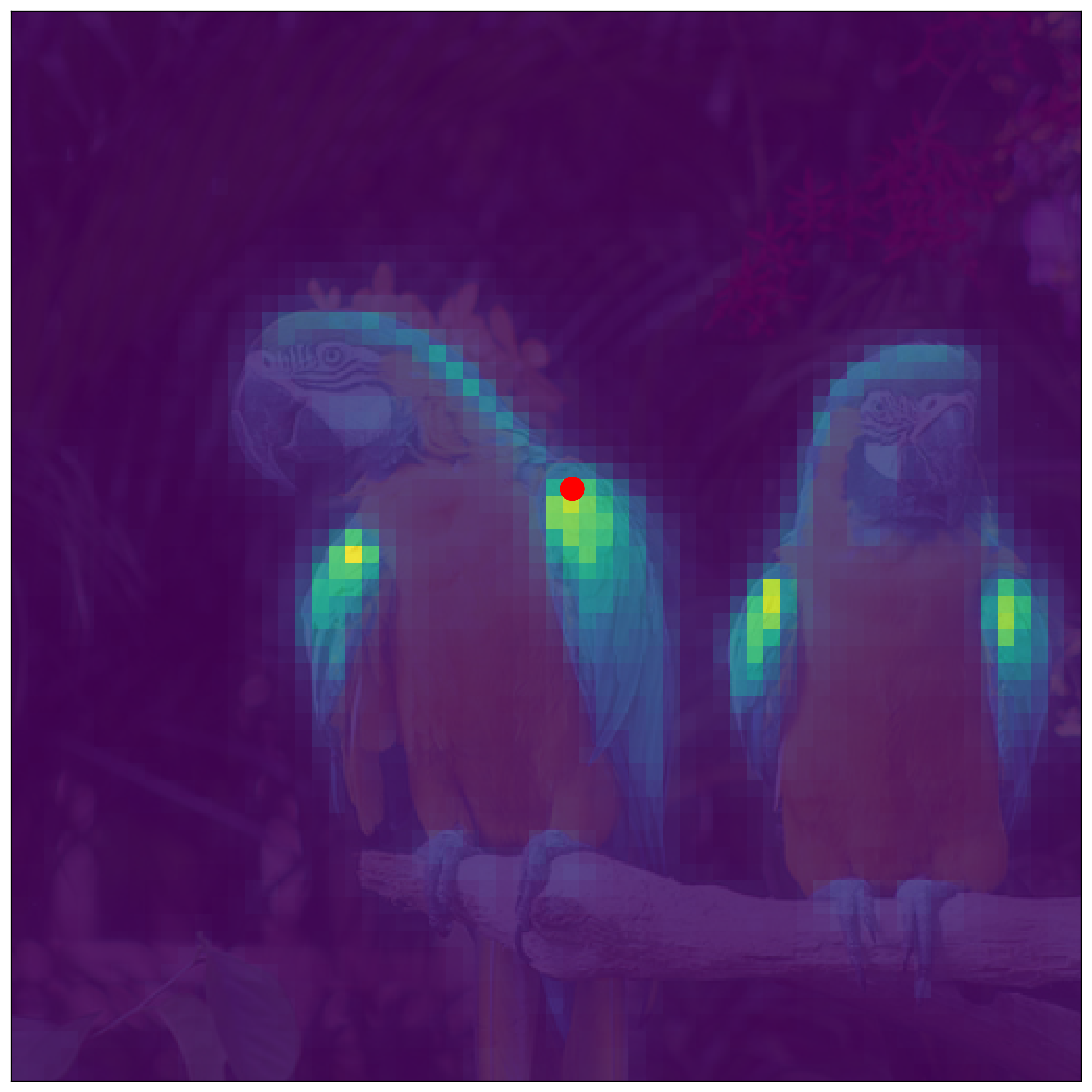}};
        \end{tikzpicture}%
        \begin{tikzpicture}
          \node[inner sep=0] (image4) {\includegraphics[height=2.5cm]{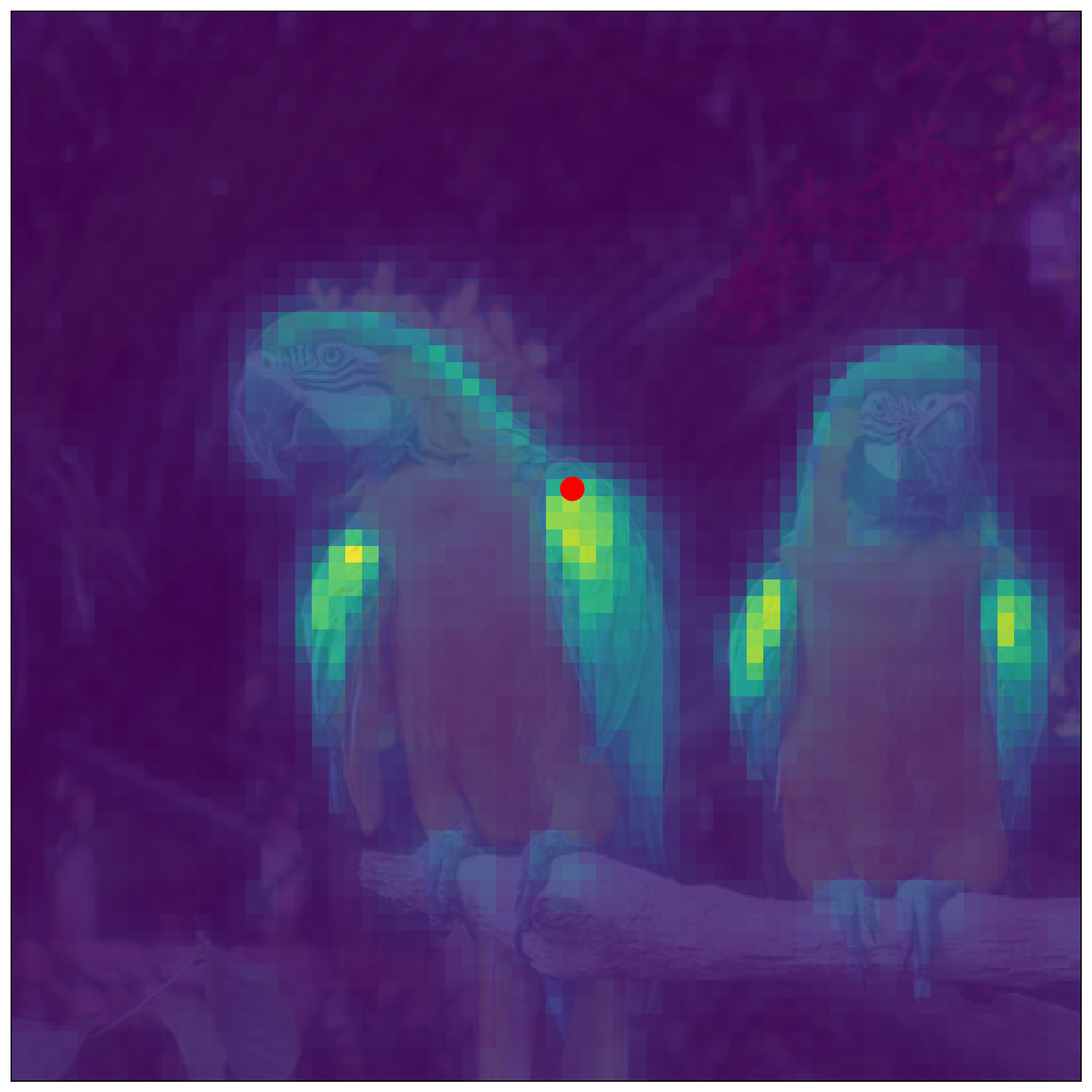}};
        \end{tikzpicture}%
        \begin{tikzpicture}
          \node[inner sep=0] (image5) {\includegraphics[height=2.5cm]{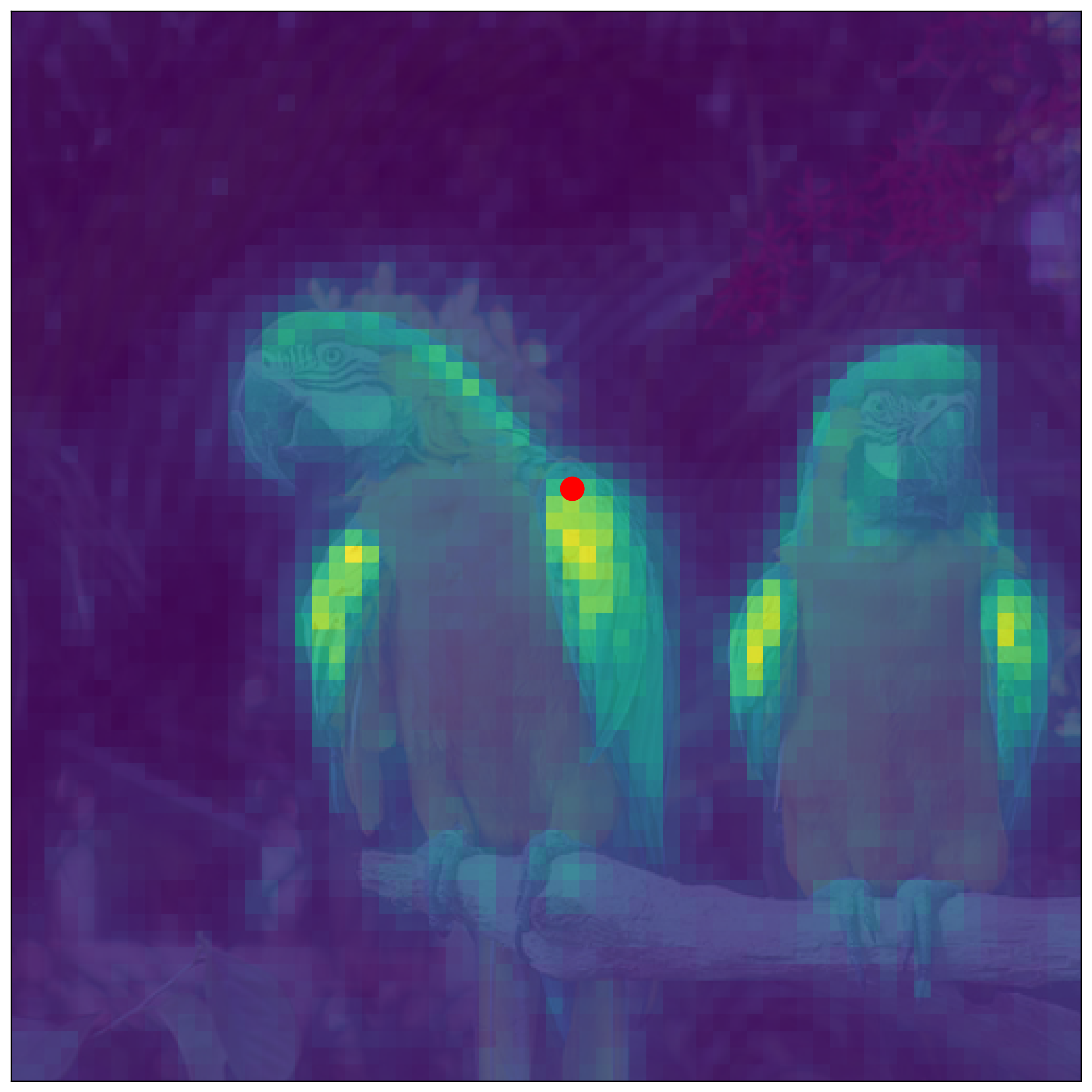}};
        \end{tikzpicture}%
    }
    \resizebox{\linewidth}{!}{%
        \begin{tikzpicture}
          \node[inner sep=0] (image6) {\includegraphics[height=2.5cm]{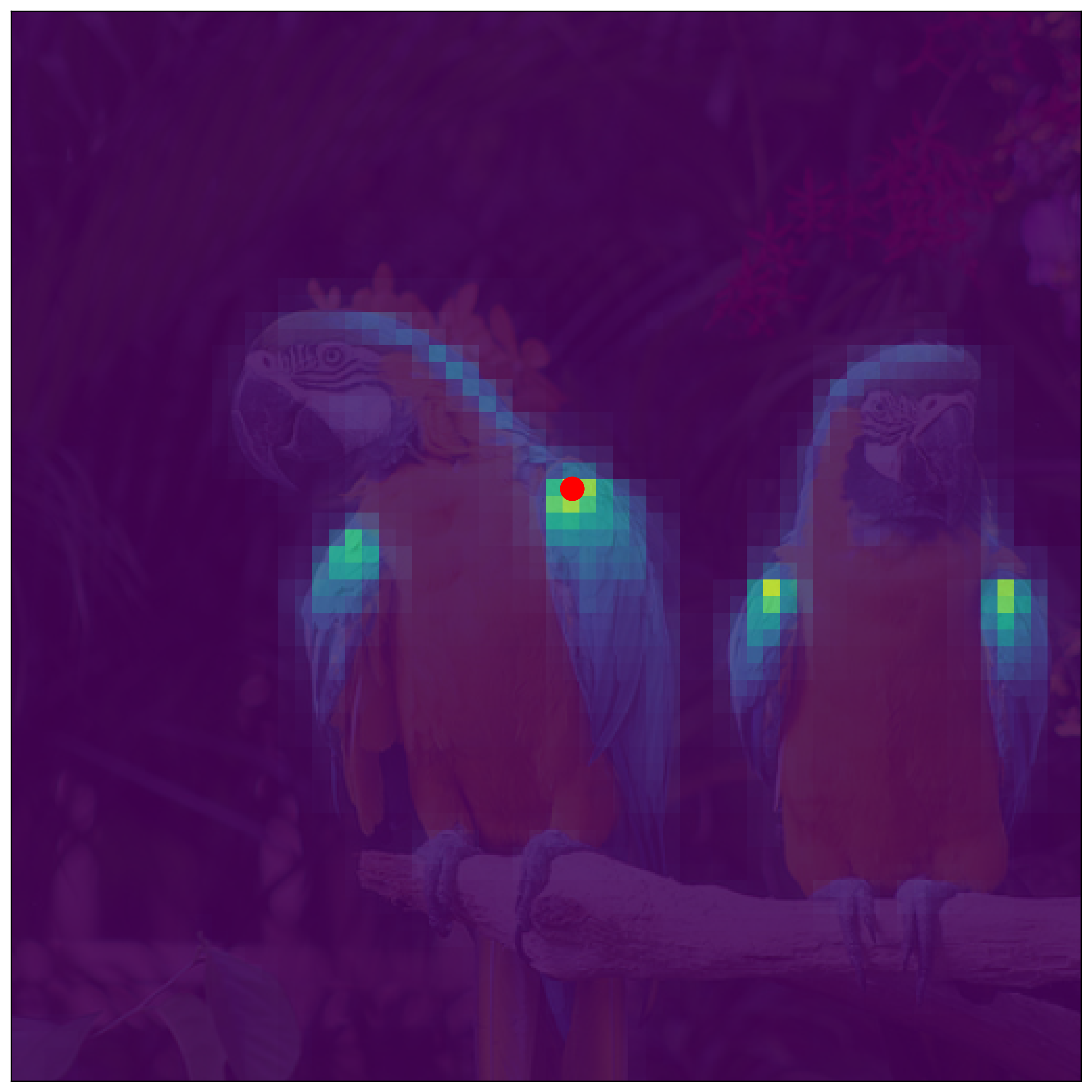}};
        \end{tikzpicture}%
        \hspace{0.1em}
        \begin{tikzpicture}
          \node[inner sep=0] (image7) {\includegraphics[height=2.5cm]{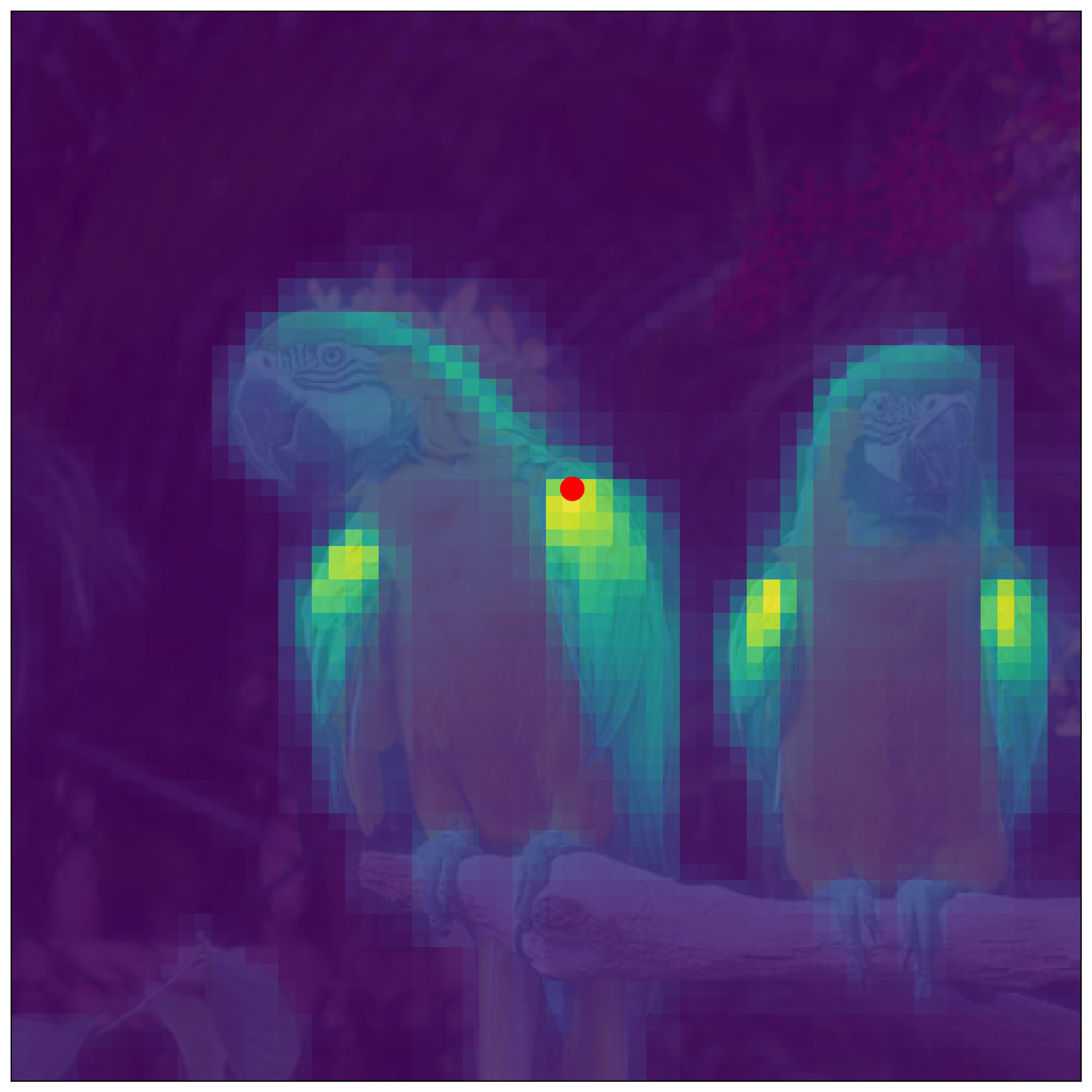}};
        \end{tikzpicture}%
        \begin{tikzpicture}
          \node[inner sep=0] (image8) {\includegraphics[height=2.5cm]{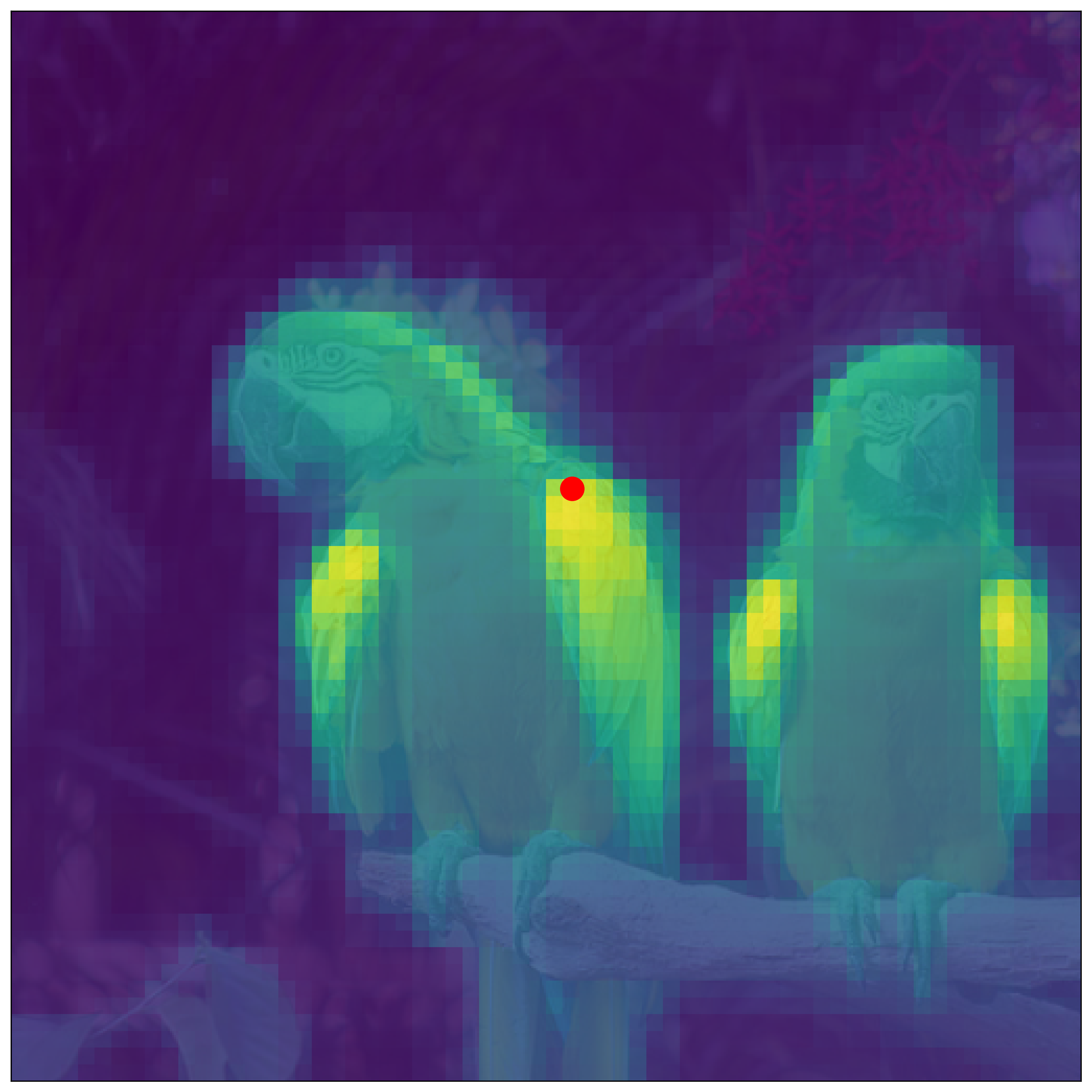}};
        \end{tikzpicture}%
        \begin{tikzpicture}
          \node[inner sep=0] (image9) {\includegraphics[height=2.5cm]{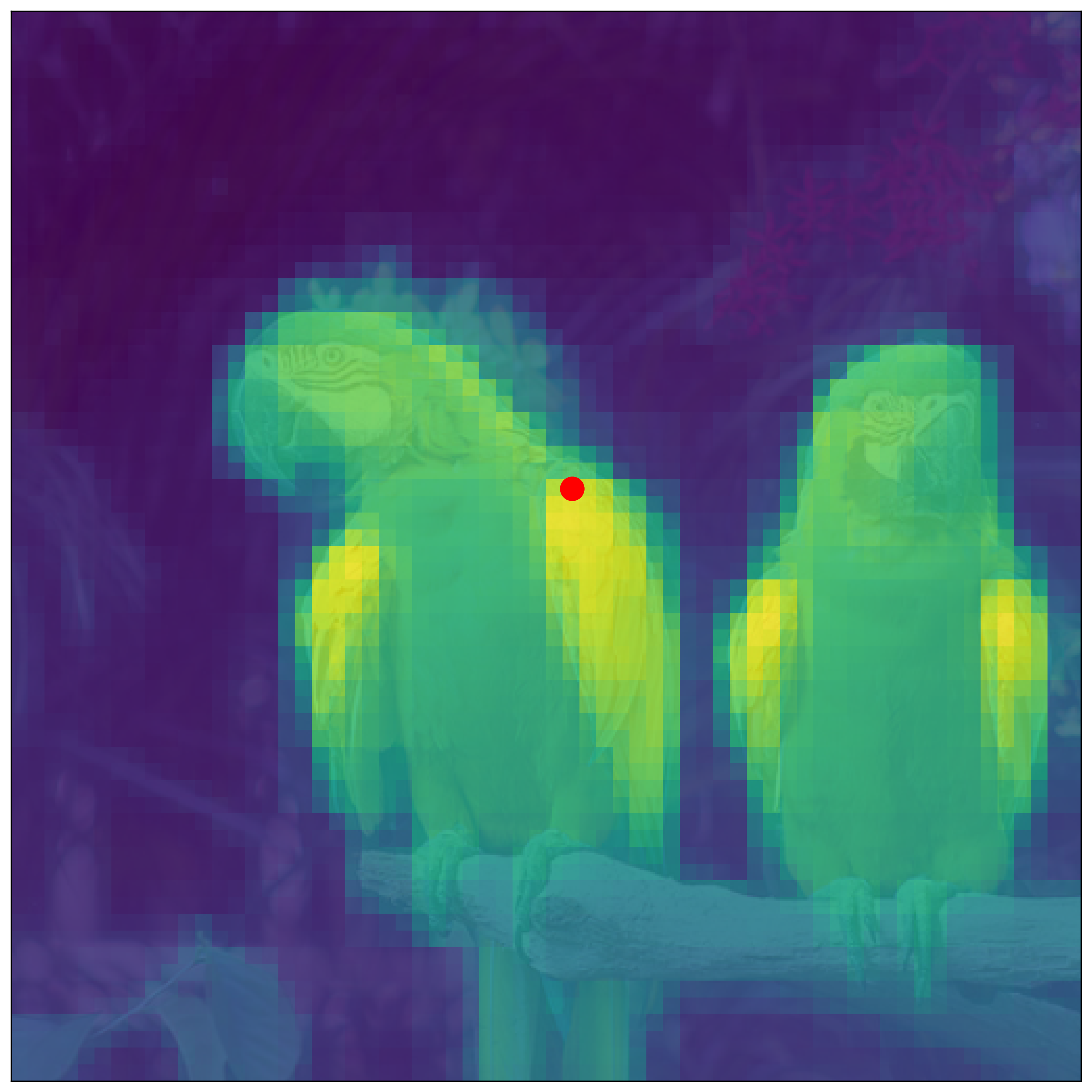}};
        \end{tikzpicture}%
        \begin{tikzpicture}
          \node[inner sep=0] (image10) {\includegraphics[height=2.5cm]{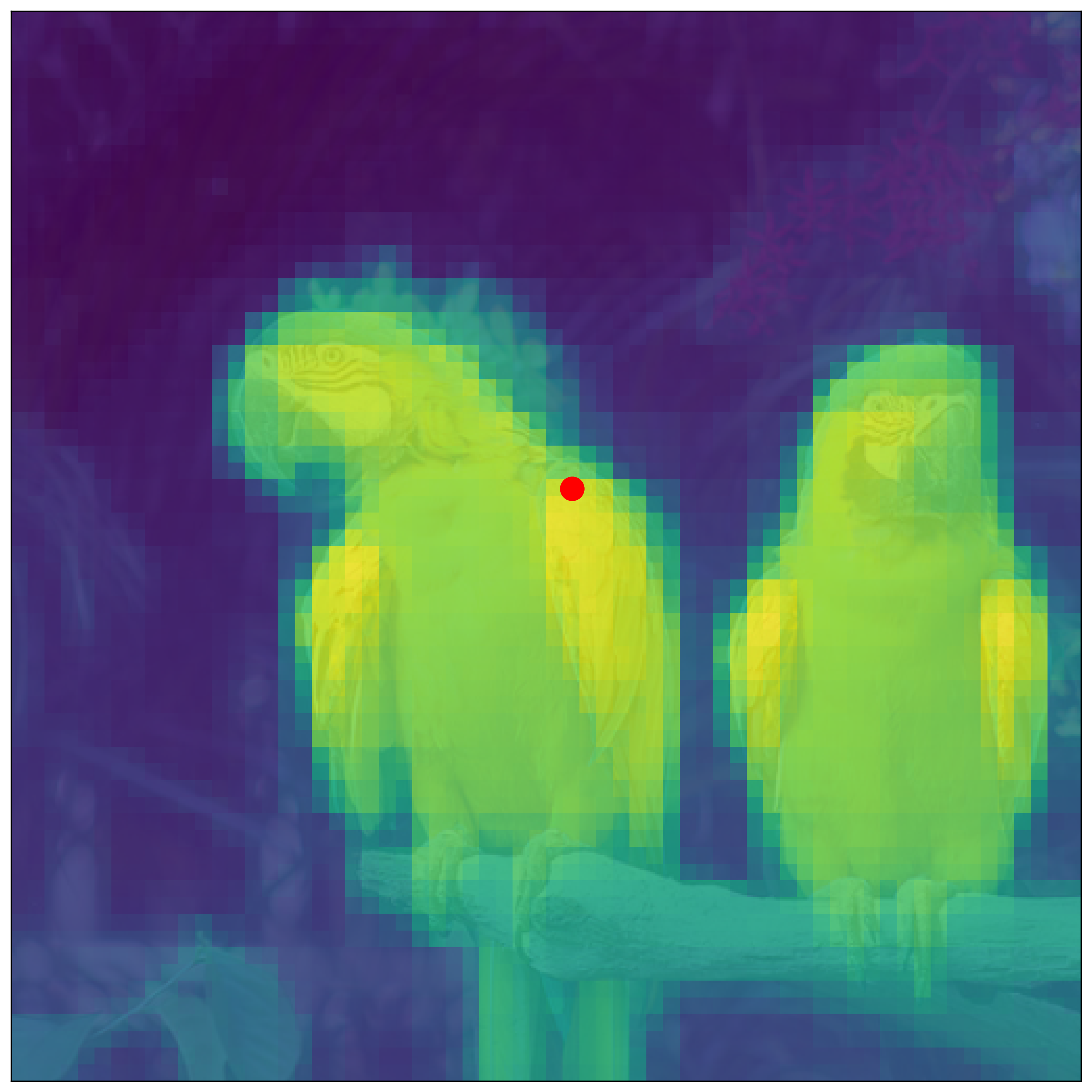}};
        \end{tikzpicture}%
    }
    \resizebox{\linewidth}{!}{%
        \begin{tikzpicture}
          \node[inner sep=0] (image11) {\includegraphics[height=2.5cm]{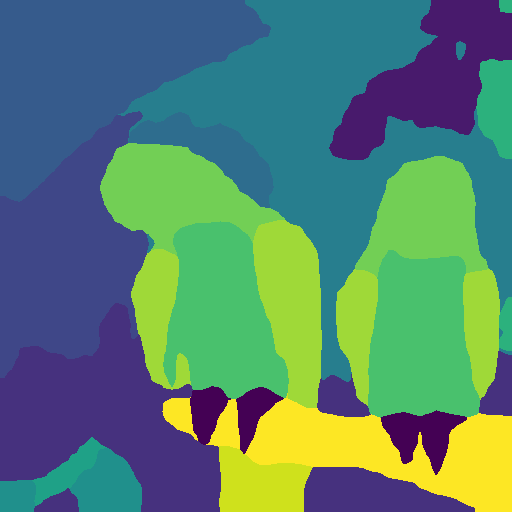}};
          \node[anchor=north] at ($(image11.south)+(0.0,0.0)$) {k=1};
        \end{tikzpicture}%
        \hspace{0.1em}
        \begin{tikzpicture}
          \node[inner sep=0] (image12) {\includegraphics[height=2.5cm]{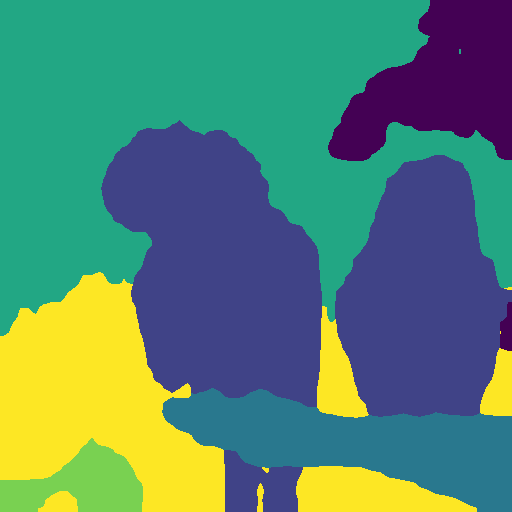}};
          \node[anchor=north] at ($(image12.south)+(0.0,0.0)$) {k=2};
        \end{tikzpicture}%
        \begin{tikzpicture}
          \node[inner sep=0] (image13) {\includegraphics[height=2.5cm]{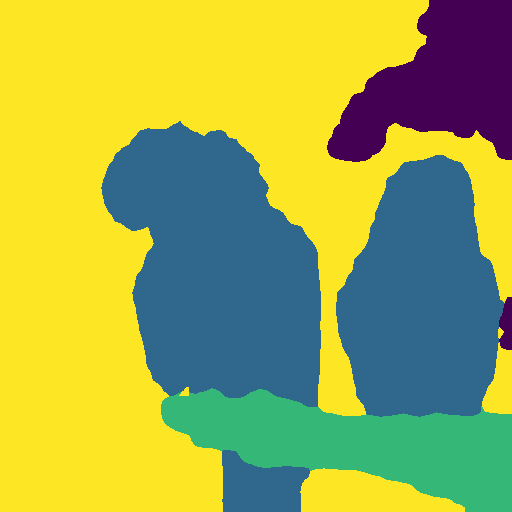}};
          \node[anchor=north] at ($(image13.south)+(0.0,0.0)$) {k=3};
        \end{tikzpicture}%
        \begin{tikzpicture}
          \node[inner sep=0] (image14) {\includegraphics[height=2.5cm]{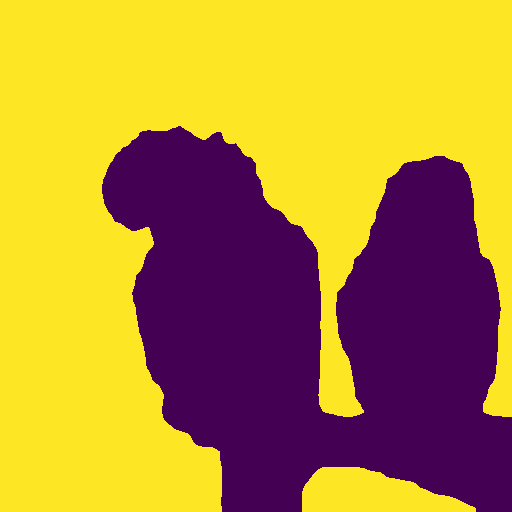}};
          \node[anchor=north] at ($(image14.south)+(0.0,0.0)$) {k=4};
        \end{tikzpicture}%
        \begin{tikzpicture}
          \node[inner sep=0] (image15) {\includegraphics[height=2.5cm]{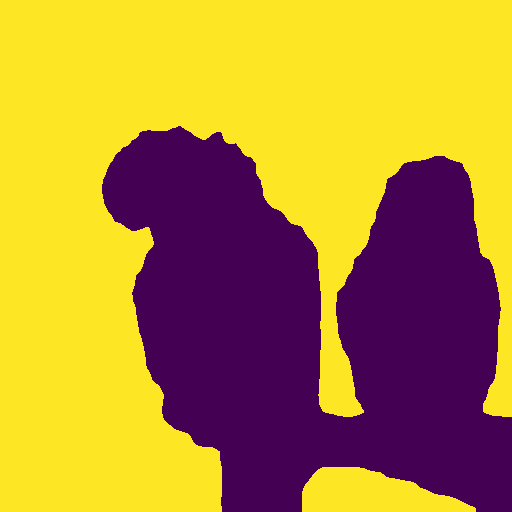}};
          \node[anchor=north] at ($(image15.south)+(0.0,0.0)$) {k=5};
        \end{tikzpicture}%
    }
\end{center}
\vspace{-12pt}
\caption{Power Walks: Self-attention extracted from Stable Diffusion for an arbitrary patch marked by red dot (row 1) framed as a random walk transition probability matrix and raised to the $k$\textsuperscript{th} power, allows us to build adjacency matrices which increasingly capture higher-order patch-similarities (row 2). The resulting unsupervised segmentations (row 3) produced by our hyperparameter-free NCut method demonstrate that as long-range relationships are enhanced, the granularity of segmentation decreases, while still extracting semantically-meaningful segments.}
\label{fig:qualitative-comparison-power-settings}
\vspace{-10pt}
\end{figure}

While our previous experiment demonstrates a fully automated approach for zero-shot unsupervised segmentation, removing the need for pre-setting the NCut threshold means we do not have a way to control the granularity of segmentations. As established in Section~\ref{subsec:method-feature-adjacency}, self-attention matrices can be interpreted as random walk transition probabilities. This allows us to modulate multi-hop relationships through matrix exponentiation prior to constructing our adjacency matrix. Our quantitative results in Table~\ref{tab:power-walks} show that considering higher order walks ($k=2,3,4$) substantially improves the performance of the Scaled MinCut threshold approach against the annotations of COCO-Stuff-27.
Qualitatively (Figure~\ref{fig:qualitative-comparison-power-settings}), we see that higher exponents have the effect of hierarchically merging the more granular segments produced at $k=1$: the heads, wings, feet and bodies of the parrots are merged into a larger "parrot" segment, which is closer to the human-annotated ground truth labels.  

\begin{table}[!h]
\small
\centering
\begin{tabularx}{\columnwidth}{@{}l*{4}{>{\centering\arraybackslash}X}*{3}{>{\centering\arraybackslash}p{0.1\columnwidth}}@{}}
\toprule
\multirow{2}{*}{\textbf{Approach}} & \multicolumn{4}{c}{\textbf{Resolution Levels}} & \multirow{2}{*}{\textbf{Acc $\uparrow$}} & \multirow{2}{*}{\textbf{F1 $\uparrow$}} & \multirow{2}{*}{\textbf{mIoU $\uparrow$}} \\
\cmidrule(lr){2-5}
 & \textbf{8} & \textbf{16} & \textbf{32} & \textbf{64} & & & \\ 
\midrule
\shortstack{Random Walk} & \checkmark & \checkmark & \checkmark & \checkmark & 72.2 & 56.9 & 43.3 \\
\midrule
\multirow{4}{*}{\shortstack{MAN-C\\ w/ NCut}} & \checkmark & \checkmark & \checkmark & \checkmark & 74.7 & 60.6 & {\ul 46.1} \\
 & & \checkmark & \checkmark & \checkmark & 74.1 & 60.8 & \textbf{46.6} \\
 & & & \checkmark & \checkmark & 70.1 & 56.4 & 42.7 \\
 & & & & \checkmark & 63.9 & 49.6 & 36.4 \\ 
\midrule
\multirow{4}{*}{\shortstack{MAN-C\\ w/ Scaled\\MinCut}} & \checkmark & \checkmark & \checkmark & \checkmark & 45.9 & 36.1 & 23.8 \\
 & & \checkmark & \checkmark & \checkmark & 42.0 & 32.0 & 20.6 \\
 & & & \checkmark & \checkmark & 37.2 & 27.6 & 17.0 \\
 & & & & \checkmark & 36.1 & 25.0 & 15.3 \\ 
\midrule
DiffSeg & \checkmark & \checkmark & \checkmark & \checkmark & 72.5 & 58.5 & 43.6 \\
\midrule
DiffCut & & & \checkmark & & 59.6 & 44.5 & 31.3 \\ 
\bottomrule
\end{tabularx}
\caption{Comparison of our automatic threshold methods across different attention resolution combinations on Coco-Stuff-27. MAN-C NCut consistently outperforms both Scaled MinCut and state-of-the-art DiffSeg. Random Walk baseline has NCut threshold set to $0.3$.}
\label{tab:man-c-comparison}
\end{table}

\subsection{Comparison with state-of-the-art}
In Table~\ref{tab:quantitative-sota} we summarise the results of our method alongside several recent approaches to unsupervised segmentation---DiffSeg~\cite{tian2024diffuse}, DiffCut~\cite{couairon2024zero} and EmerDiff~\cite{namekata2024emerdiff}, which are the closest to it conceptually, being zero-shot unsupervised segmentation models which rely on Stable Diffusion features and represent state-of-the-art. \revise{We also report results from \emph{trained} unsupervised state-of-the-art methods PiCIE~\cite{cho2021picie}, STEGO~\cite{hamilton2022unsupervised}, DepthG~\cite{Sick_2024_CVPR}, U2Seg~\cite{Niu_2024_CVPR}, CUPS~\cite{Hahn:2025:UPS} and EAGLE~\cite{2024eagle}.}

\begin{table*}[!h]
\centering
\resizebox{\textwidth}{!}{%
\begin{tabular}{@{}lcccccccccc@{}}
\toprule
 & \multicolumn{6}{c}{\textbf{Trained}} & \multicolumn{4}{c}{\textbf{Zero-shot}} \\
  & \textbf{PiCIE}~\cite{cho2021picie} & \textbf{STEGO}~\cite{hamilton2022unsupervised} & \textbf{\revise{DepthG}}~\cite{Sick_2024_CVPR} & \textbf{\revise{U2Seg}}~\cite{Niu_2024_CVPR} & \textbf{\revise{CUPS}}~\cite{Hahn:2025:UPS} & \textbf{\revise{EAGLE}}~\cite{2024eagle} & \textbf{EmerDiff}~\cite{namekata2024emerdiff} & \textbf{DiffSeg}~\cite{tian2024diffuse} & \multicolumn{1}{l}{\textbf{DiffCut}}~\cite{couairon2024zero} & \textbf{Ours} \\ \midrule
COCO-Stuff & 13.8 & 28.2 & 29.0 & 30.2 & - & 27.2 & 26.6 & {\ul 43.6} & 31.3 & \textbf{46.6} \\
Cityscapes & 12.3 & 21.0 & 23.1 & 21.6 & 26.8 & 22.1 & 12.0 & {\ul 21.2} & 18.1 & \textbf{22.0} \\
ADE20K & - & - & - & - & - & - & - & 37.7 & {\ul 40.1} & \textbf{41.2} \\\bottomrule
\end{tabular}%
}
\vspace{-5pt}
\caption{Quantitative comparison of mIoU between our approach and current unsupervised zero-shot state-of-the-art on COCO-Stuff-27, Cityscapes and ADE20K, \revise{along with non-zero-shot unsupervised state-of-the-art for completeness}. Best in bold, second best underscored.}
\vspace{-10pt}
\label{tab:quantitative-sota}
\end{table*}

For DiffSeg, we use their suggested best hyperparameters, i.e.~$1.1$ and $0.9$ KLD threshold for COCO-Stuff and Cityscapes respectively, and report additional ADE20K results from ~\cite{couairon2024zero}. For EmerDiff, we evaluate with Hungarian matching (dubbed ``traditional evaluation strategy" by the authors) for a fair comparison against the other methods, and set the mask proposal parameter $K=30$, as the authors suggest. While both DiffSeg and EmerDiff rely on Stable Diffusion 1.4 as a feature extractor, DiffCut report their main results on SSD 1B~\cite{gupta2024progressive}, a distilled variant of Stable Diffusion XL\cite{podellsdxl}, and utilise a post-processing step to refine segmentation predictions - PAMR~\cite{araslanov2020single}. For fairness, we exclude PAMR, and evaluate the method on Stable Diffusion 1.4 using hyperparameters $\tau=0.2$ and $\alpha=10$, optimised for SD1.4 by the authors in their ablation. 
We include the best-performing variant of our method, which uses three self-attention resolutions, and moreover does not rely on any hyperparameters as it leverages our dynamic NCut thresholding scheme; this variant improves on the state-of-the-art DiffSeg by 3.0\% mIoU on COCO-Stuff and 0.8\% on Cityscapes. In fact, some of the weaker variants of our approach still outperform DiffSeg, see Tables~\ref{tab:adjacency-dot-vs-cos}~\&~\ref{tab:man-c-comparison}. Furthermore, our dynamic threshold approach outperforms DiffCut, despite it using NCut hyperparameters specifically optimised for Stable Diffusion 1.4. In terms of computational cost, all variants take 2-4s per image, with the highest-scoring one (w.r.t. quantitative metrics) taking 2.5s on an A5000 GPU and using 16GB of VRAM.

\begin{table}[h]
\centering
\small
\begin{tabular}{@{}c c c c@{}}
\toprule
\textbf{Model} & \textbf{Method} & \textbf{w/o PAMR} & \textbf{w/ PAMR} \\
\midrule
\multirow{2}{*}{SSD 1B} 
 & DiffCut & 46.1 & 49.1 \\
 & Ours    & 41.3 & 42.6 \\
\midrule
\multirow{2}{*}{SD 1.4}
 & DiffCut & 31.3 & 45.2 \\
 & Ours    & 46.6 & 47.1 \\
\bottomrule
\end{tabular}
\caption{Comparison with DiffCut (mIoU) on COCO-Stuff-27.}
\label{tab:diffcut_comparison}
\end{table}
We perform further ablation against DiffCut Table~\ref{tab:diffcut_comparison}. For fair comparison, we benchmark our approach against DiffCut using both SD 1.4 and SSD 1B. We also ablate the effect PAMR postprocessing by adding it to our pipeline as well. We use the best hyperparameters suggested by the authors, i.e. $\mathrm{\alpha}=10$ and $\mathrm{\tau}=0.5$ for SSD 1B or $\mathrm{\tau}=0.2$ for SD 1.4. 

With SD 1.4, our method achieves $46.6$ mIoU on Coco-Stuff-27 without PAMR post-processing outperforming DiffCut's $45.2$ mIoU \emph{with} PAMR and $31.3$ without. In fact, without PAMR, our SD 1.4 results ($46.6$ mIoU) outperform DiffCut even with the newer, larger SSD 1B model. 

Unlike SD 1.4, SSD 1B lacks self-attention at $8\times8$ and $16\times16$ resolutions, and has an asymmetric encoder-decoder structure with additional decoder transformer blocks (34 in total), hence the DiffSeg feature aggregation strategy does not translate to this setting. While our initial results with this backbone show promise (Table~\ref{tab:diffcut_comparison}), optimization of feature aggregation strategies for SSD 1B might improve performance.  

Qualitatively, both our coarser and finer predictions (Figure~\ref{fig:qualitative-small}) are more precise than those of DiffSeg and DiffCut, with additional visual comparisons in the Supplementary.

\section{Conclusions}
\label{sec:conclusions}
Unsupervised image segmentation is a challenging and ambiguous task where even the definition of ``correct" remains debatable~\cite{eramian2020benchmarking}. 
We demonstrate that we can exploit the naturally emerging patterns in diffusion self-attentions to discover meaningful segmentations, by exploring the activations through the lens of random walks. By framing self-attention as transition probabilities in a Markov process, we establish a principled approach for clustering image regions based on their semantic relationships, while offering intuitive control over segmentation granularity through walk length. Our approach combines the strengths of both top-down spectral clustering methods (like DiffCut) and bottom-up merging strategies (like DiffSeg) while addressing their limitations: we eliminate the need for manual threshold tuning in recursive partitioning through our automatic threshold determination, while avoiding the grid sampling and iterative merging inefficiencies of bottom-up approaches through our power walk exponentiation, which naturally handles the ``merging" of semantically related regions. This balanced methodology produces hierarchically consistent segmentations that align with human perceptual judgments across diverse image types.

Our approach reveals intrinsic object groupings in learned representations, specifically self-attention layers, while achieving superior quantitative results. Though implemented for Stable Diffusion 1.4, our methodology is model agnostic and applicable to any model with attention features, as demonstrated with SSD 1B in our ablation against DiffCut. 

Variants that most closely matched human annotation granularity naturally produced the best quantitative results, while qualitative analysis reveals that even lower-scoring variants maintain consistent semantic groupings. Future work should investigate the discrepancies between human annotations and features learned by foundation models.
\section*{Acknowledgements}
This work was supported by the Engineering and Physical Sciences Research Council [grant number EP/R513222/1].
{
    \small
    \bibliographystyle{ieeenat_fullname}
    \bibliography{main}
}

\clearpage 
\clearpage
\setcounter{page}{1}
\maketitlesupplementary
\appendix

\section{A note on evaluation strategies}

Zero-shot unsupervised segmentation presents unique evaluation challenges compared to traditional supervised approaches. This section discusses three complementary evaluation strategies that better capture different aspects of zero-shot performance.
\paragraph{Traditional Global Averaging vs. Per-Image Evaluation}
While previous non-zero-shot methods typically use global averaging (accumulating statistics across all images before calculating metrics)~\cite{cho2021picie, hamilton2022unsupervised}, this may not be ideal for zero-shot models. Global averaging emphasizes performance on frequent classes, but zero-shot models have no prior knowledge of class distributions. Instead, per-image averaging, which computes metrics independently for each image before averaging, treats each instance of class discovery equally - better reflecting the zero-shot nature of the task.

\paragraph{The Challenge of Granularity}
A key challenge in evaluating zero-shot models is their tendency to discover more fine-grained segmentations than present in the ground truth. For example, in Figure~\ref{fig:qualitative-comparison-segmentation-coco}, row 7, where the ground truth groups a bear and bird into a single "animal" class, models trained on the CocoStuff dataset e.g. STEGO~\cite{hamilton2022unsupervised} also group them together; in contrast, our model distinguishes them as separate classes. While this more detailed segmentation may be semantically meaningful, it is penalized by both global and per-image metrics.

\begin{figure*}[!htpb]
\begin{center}
    \includegraphics{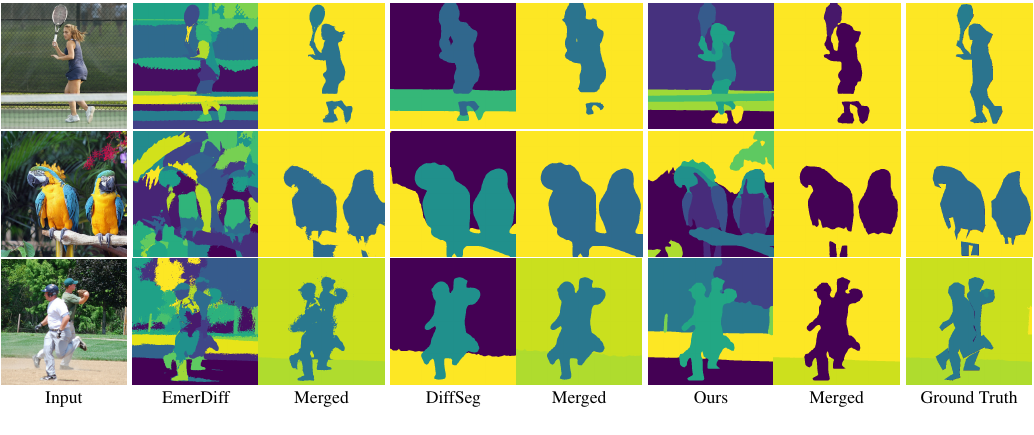}
\end{center}
\caption{Examples of predictions used in the traditional evaluation setting (Hungarian matching) from the zero-shot diffusion feature-based approaches (columns 2, 4, and 6), and the corresponding "merged" predictions (columns 3, 5, and 7), designed to account for the models segmenting more granular classes compared to what is given in the ground truth. Notice that predictions from our model can be even more precise (baseball players example, row 3) or more accurate (tennis net covering the player's legs, row 1).}
\label{fig:merging}
\end{figure*}

\paragraph{Oracle-Merged Evaluation}
To address this, we introduce an oracle-merged evaluation strategy that uses ground truth to merge oversegmented areas based on their primary class overlap. This is conceptually similar to EmerDiff's approach~\cite{namekata2024emerdiff}, though they merge regions based on embedding similarity. However, EmerDiff requires pre-specifying the number of clusters (K=30), which leads to extreme oversegmentation as shown in Figure~\ref{fig:merging}. While this hurts their performance under standard metrics, they benefit significantly from merged evaluation strategies, as discussed below.

Our extended evaluation framework combines all three approaches. As shown in Table~\ref{tab:extended-metrics}, our method outperforms other diffusion-based zero-shot baselines across all metrics. EmerDiff achieves higher accuracy only in the oracle-merged setting, likely due to their cross-attention modulating upsampling approach. The qualitative examples in Figure~\ref{fig:merging} demonstrate why our approach still outperforms EmerDiff in the more strict F1 and mIoU metrics - despite benefiting from a more sophisticated upsampling strategy, EmerDiff produces in noisy clusters. Furthermore, even our "oversegmented" versions produce high-quality, semantically meaningful segments, without the need to specify the number of clusters or any other hyperparameter a priori, unlike EmerDiff and DiffSeg~\cite{tian2024diffuse}.

Importantly, this extended evaluation framework further underscores that metrics don't necessarily measure segmentation "correctness" (which is inherently subjective), but rather alignment with human-labeled ground truths - which can vary in granularity across datasets and images.

\begin{table}[!hbtp]
\centering
\resizebox{\columnwidth}{!}{%
\begin{tabular}{@{}llllllllll@{}}
\toprule
 & \multicolumn{3}{l}{\begin{tabular}[c]{@{}l@{}}Globally averaged \\ (traditional) \end{tabular}} & \multicolumn{3}{l}{\begin{tabular}[c]{@{}l@{}}Per-image\\ averaged\end{tabular}} & \multicolumn{3}{l}{\begin{tabular}[c]{@{}l@{}} Merged and \\ per-image\\ averaged \end{tabular}} \\ \midrule
 & Acc & F1 & mIoU & Acc & F1 & mIoU & Acc & F1 & mIoU \\ \midrule
\revise{STEGO} & 56.9 & - & 28.2 & 73.3 & - & 23.3 & - & - & - \\
EmerDiff & 28.8 & 22.8 & 13.3 & 28.9 & 47.8 & 0.1 & 92.5 & 34.9 & 24.0 \\
DiffSeg & 72.5 & 58.5 & 43.6 & 72.6 & 71.1 & 34.1 & 79.2 & 73.0 & 60.6 \\
Ours & 74.1 & 60.8 & 46.6 & 74.0 & 73.3 & 38.0 & 83.6 & 76.0 & 64.5 \\ \bottomrule
\end{tabular}%
}
\caption{We calculate metrics on predictions from Coco-Stuff-27 validation set under three different settings: globally averaged, per-image averaged, and per-image averaged with an oracle which merges oversegmented regions; our approach excels in all three.}
\label{tab:extended-metrics}
\end{table}
\vspace{-15pt}

\section{Additional qualitative examples}
\begin{figure*}[!htpb]
\begin{center}
    \includegraphics{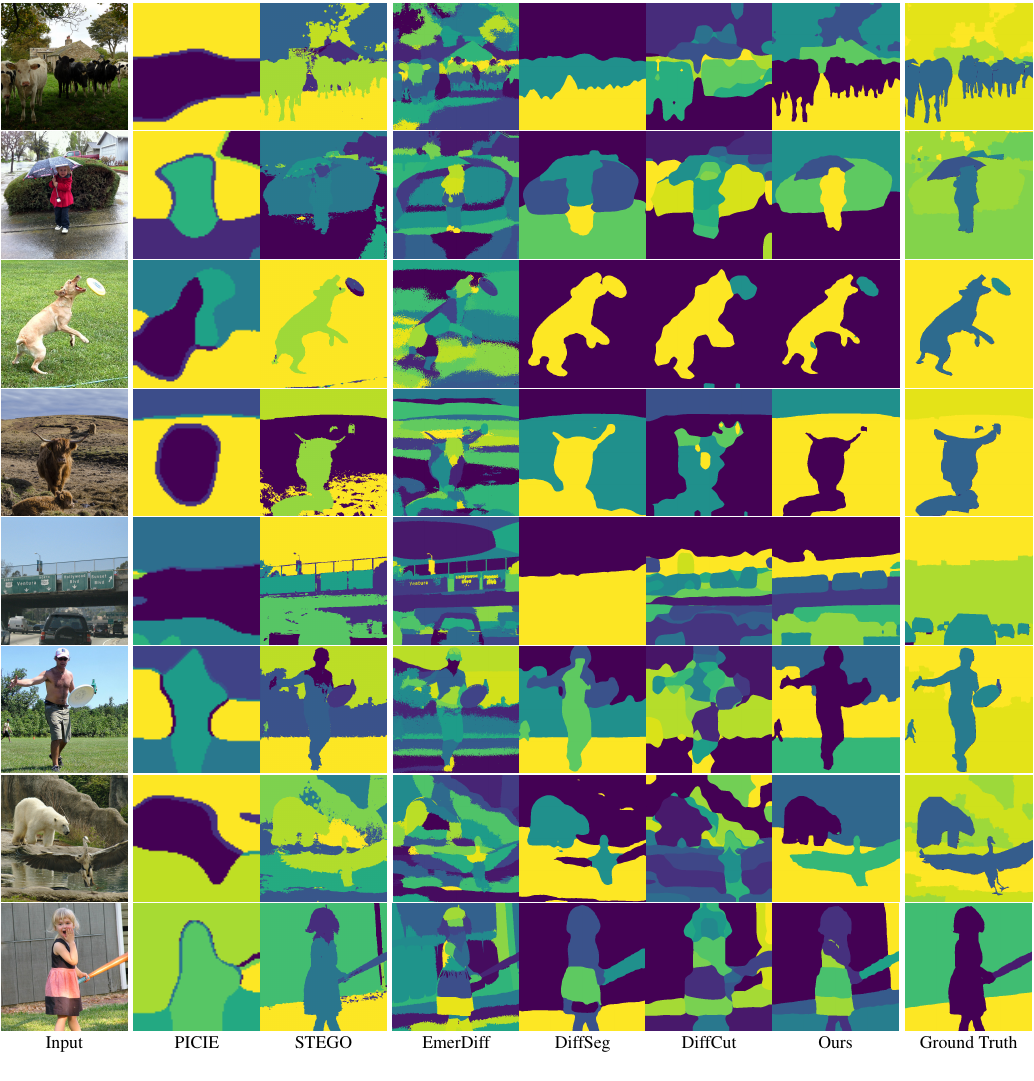}
\end{center}
\caption{Qualitative comparison of segmentation results on Coco-Stuff-27~\cite{caesar2018coco} Validation across multiple methods. All approaches are unsupervised; PICIE and STEGO are fit on the training set of Coco-Stuff-27, while EmerDiff, DiffSeg, DiffCut and Ours are all zero-shot and rely only on features extracted from Stable Diffusion. Our approach, in contrast to DiffSeg, DiffCut and EmerDiff, is also hyperparameter-free.}
\label{fig:qualitative-comparison-segmentation-coco}
\end{figure*}

\begin{figure*}[!htpb]
\begin{center}
    \includegraphics{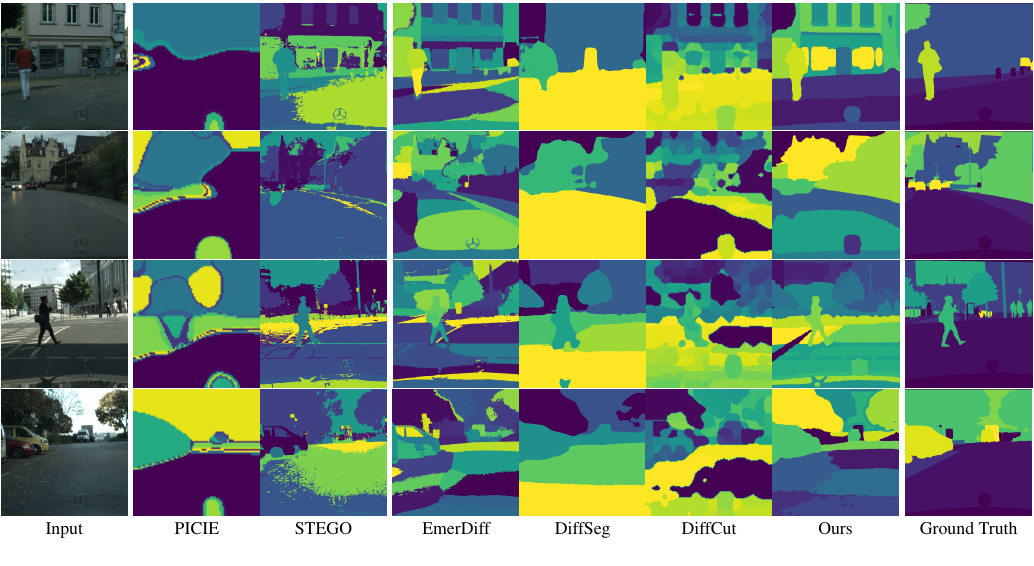}
\end{center}
\caption{Qualitative comparison of segmentation results on Cityscapes~\cite{Cordts2016Cityscapes} Validation across multiple methods. All approaches are unsupervised; PICIE and STEGO are fit on the training set of Cityscapes, while EmerDiff, DiffSeg, DiffCut and Ours are all zero-shot and rely only on features extracted from Stable Diffusion. Our approach, in contrast to DiffSeg, DiffCut and EmerDiff, is also hyperparameter-free.}
\label{fig:qualitative-comparison-segmentation-city}
\end{figure*}

\begin{figure*}[!htpb]
\begin{center}
    \includegraphics{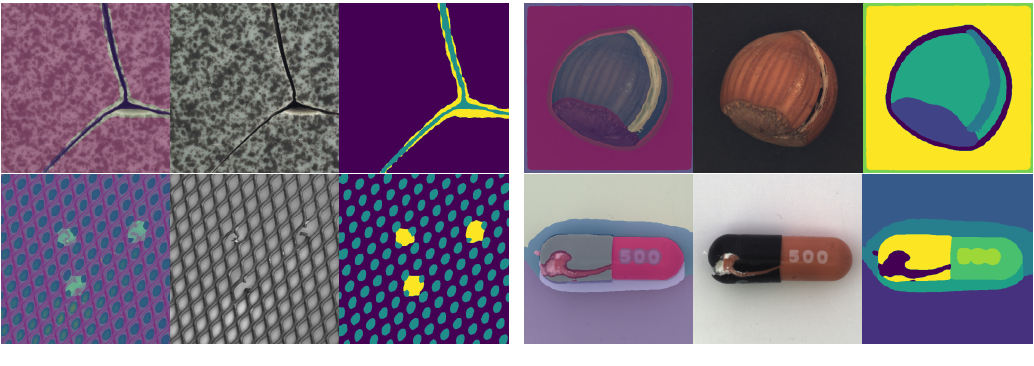}
\end{center}
\caption{Segmentations of various types of anomalous objects from the MVTec Dataset~\cite{Bergmann_2019_CVPR}.}
\label{fig:mvtec}
\end{figure*}

\begin{figure*}[!htpb]
\begin{center}
    \includegraphics{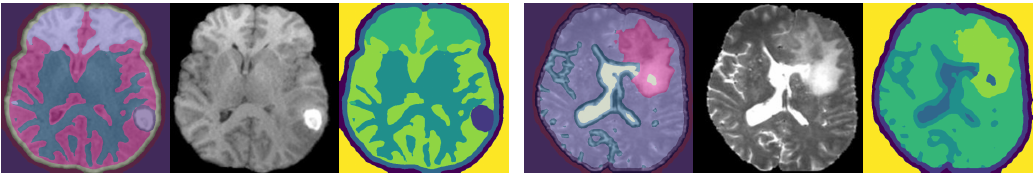}
\end{center}
\caption{Segmentations of Brain MRI scans from the BraTS 2017 Dataset~\cite{brats}.}
\label{fig:mri}
\end{figure*}

\begin{figure*}[!htpb]
\begin{center}
    \includegraphics{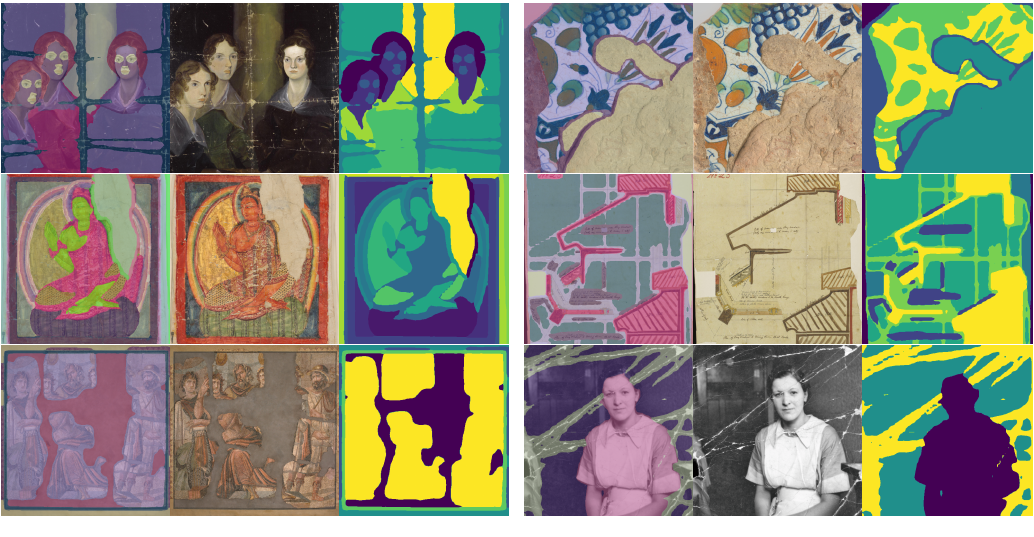}
\end{center}
\caption{Segmentations of varying granularities on different types of damaged analogue media~\cite{ivanova2024state}.}
\label{fig:analogue-media}
\end{figure*}

\begin{figure*}[!htpb]
\begin{center}
    \includegraphics{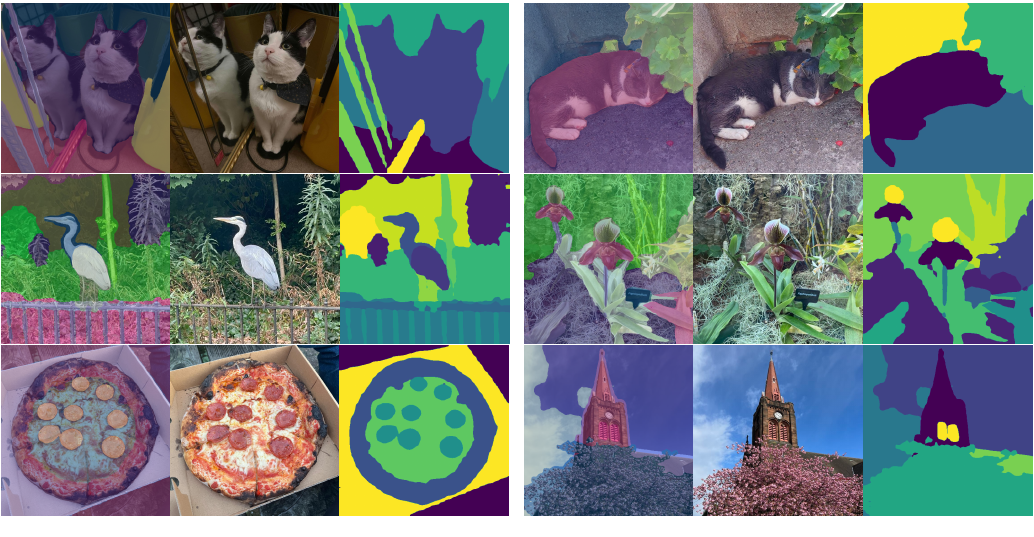}
\end{center}
\caption{Segmentations of varying granularities of in-the-wild images captured by a smartphone: overlay (left), input (middle), and segmentation (right).}
\label{fig:in-the-wild}
\end{figure*}

Besides qualitative comparison for all methods on the two standard segmentation benchmark datasets CocoStuff-27~\cite{caesar2018coco} in Figure~\ref{fig:qualitative-comparison-segmentation-coco} and Cityscapes~\cite{Cordts2016Cityscapes} in Figure~\ref{fig:qualitative-comparison-segmentation-city}, we also visualize segmentations obtained via our proposed method on anomalous objects from MVTec~\cite{Bergmann_2019_CVPR} in Figure~\ref{fig:mvtec}; Brain MRI scans from BraTS 2017~\cite{brats} in Figure~\ref{fig:mri}, damaged analogue media~\cite{ivanova2024state} in Figure~\ref{fig:analogue-media} and real-life images captured in the wild in Figure~\ref{fig:in-the-wild}.

\section{Ablation on feature extraction timestep}
Regarding DDIM inversion and choice of noising step, we conduct ablation experiments, shown in Table~\ref{tab:ddim-ablation}. Our method performs robustly across different timesteps [30,45], and maintains performance even without DDIM inversion. 
\begin{table}[h]
\centering
\tiny
\resizebox{\linewidth}{!}{%
\begin{tabular}{@{}lcccccc@{}}
\toprule
Step & 25 & 30 & 35 & 40 & 45 & 40 no inv. \\
\midrule
mIoU & 43.1 & 45.7 & 46.6 & 46.6 & 45.3 & 46.6 \\
\bottomrule
\end{tabular}
}
\caption{Ablation over noising step and importance of DDIM Inversion (mIoU) on Coco-Stuff-27.}
\vspace{-10pt}
\label{tab:ddim-ablation}
\end{table}
\vspace{-10pt}


\end{document}